\documentclass[12pt]{report}
\usepackage[utf8]{inputenc}
\usepackage[T1]{fontenc}
\usepackage[english]{babel}  
\usepackage[a4paper, includeheadfoot, top=0.5in, bottom=0.5in, left=1.25in, right=1.25in, headheight=15pt]{geometry}

\def\Plus{\texttt{+}}

\usepackage{caption}
\usepackage{float}
\usepackage{amsmath}
\usepackage[usenames,dvipsnames,svgnames,table]{xcolor}
\usepackage{array}
\usepackage{enumitem}

\usepackage{fancyhdr}  
\fancypagestyle{style1}{
\fancyhf{}
\fancyhead[RE,LO]{\textsc{References}}
\fancyhead[LE,RO]{Mor\'{o}n Hern\'{a}ndez 2017}
\fancyfoot[C]{\thepage}
}

\fancypagestyle{style2}{
\fancyhf{}
\fancyhead[RE,LO]{\textsc{Appendices}}
\fancyhead[LE,RO]{Mor\'{o}n Hern\'{a}ndez 2017}
\fancyfoot[C]{\thepage}
}

\usepackage{graphicx}
\graphicspath{ {images/} }
\usepackage{eso-pic}

\makeatletter  
\def\thebibliography#1{\chapter*{References\@mkboth
  {REFERENCES}{REFERENCES}}\list
  {[\arabic{enumi}]}{\settowidth\labelwidth{[#1]}\leftmargin\labelwidth
    \advance\leftmargin\labelsep
    \usecounter{enumi}}
    \def\newblock{\hskip .11em plus .33em minus .07em}
    \sloppy\clubpenalty4000\widowpenalty4000
    \sfcode`\.=1000\relax}
\makeatother

\makeatletter
\renewenvironment{thebibliography}[1]
      {\section*{\refname}%
       \@mkboth{\MakeUppercase\refname}{\MakeUppercase\refname}%
       \list{\@biblabel{\@arabic\c@enumiv}}%
            {\settowidth\labelwidth{\@biblabel{#1}}%
             \leftmargin\labelwidth
             \advance\leftmargin20pt
             \advance\leftmargin\labelsep
             \setlength\itemindent{-20pt}
             \@openbib@code
             \usecounter{enumiv}%
             \let\p@enumiv\@empty
             \renewcommand\theenumiv{\@arabic\c@enumiv}}%
       \sloppy
       \clubpenalty4000
       \@clubpenalty \clubpenalty
       \widowpenalty4000%
       \sfcode`\.\@m}
      {\def\@noitemerr
        {\@latex@warning{Empty `thebibliography' environment}}%
       \endlist}
\renewcommand\newblock{\hskip .11em\@plus.33em\@minus.07em}
\makeatother

\emergencystretch=\maxdimen
\title{\Large Paraphrasing verbal metonymy through \\computational methods. }
\author{Alberto Mor\'on Hern\'andez}
\vspace{15mm}
\date{\parbox{\linewidth}{\centering2017\endgraf
  \vspace{25mm}
  A dissertation submitted to The University of Manchester for the degree of Bachelor of Arts in the Faculty of Humanities.
}
\vfill
{\centering
  Supervisor: Dr. Andrew Koontz-Garboden\\
  \vspace{5mm}
  School of Arts, Languages and Cultures
  \vspace{15mm}
}}

\begin{document}
\pagenumbering{gobble}

\AddToShipoutPictureBG*{%
  \AtPageUpperLeft{\hspace{30mm}\raisebox{-\height}{\includegraphics[width=5.5cm]{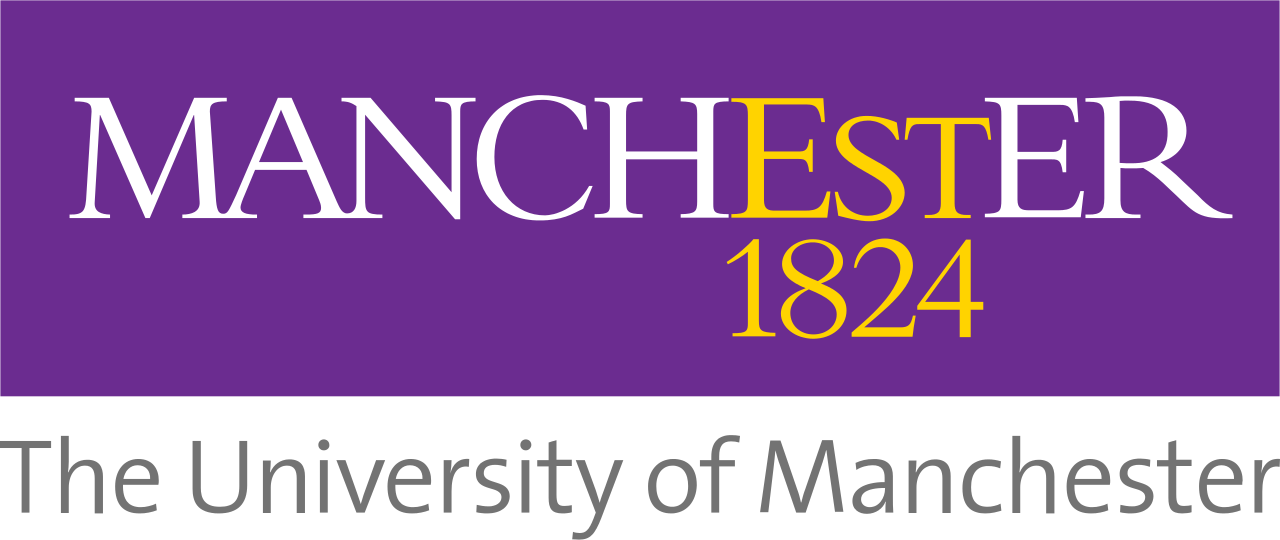}}}}
\maketitle

\newpage\thispagestyle{plain}  
\pagenumbering{roman}
I have read and understood the University of Manchester guidelines on plagiarism and declare
that this dissertation is all my own work except where I indicate otherwise by proper use of
quotes and references.

\newpage\thispagestyle{plain}  
\section*{Acknowledgements}

\hfill\vspace{2mm}
\begin{table}[H]
\centering
\begin{tabular}{c}
\textit{“Prefiero caminar con una duda que con un mal axioma.”}\hfill\break\\
\hfill---Javier Krahe
\end{tabular}
\end{table}

Thousands of words and hundreds of lines of code do not write themselves, so there are naturally many people who have my gratitude:\hfill\break

My parents, first and foremost, for the unconditional support they have given me during my time away from home. My family, both in Spain and the UK, for always being there for me.\hfill\break

Lecturers at Manchester who have guided me during the past three years, including but not limited to: Dr. Andrew Koontz-Garboden, Dr. Laurel MacKenzie, Dr. Eva Schultze-Berndt and Dr. Wendell Kimper.\hfill\break

This dissertation would never have been possible without my time at UMass Amherst. Prof. Brian Dillon, Alan Zaffeti, everyone in LING492B and Jamie, Grusha, Amy \& Ben have my gratitude for welcoming me to UMass and making it such a memorable experience.\hfill\break

Finally -- Ra\'ul, DJ, Serguei \& Pedro. Gracias. \textit{La pr\'oxima, en Cazorla.}

\newpage\thispagestyle{plain}  
\section*{Disclaimers}
This dissertation is supported in part by an Amazon Web Services Educate grant
(\#PC1R88EPEV238VD). Any opinions, findings, and conclusions or recommendations
expressed in this material are those of the author and do not necessarily reflect the view of
Amazon.com, Inc.
\hfill\break

Data cited herein has been extracted from the British National Corpus, managed by Oxford
University Computing Services on behalf of the BNC Consortium. All rights in the texts cited
are reserved.

\newpage
\tableofcontents

\newpage
\listoffigures

\newpage
\listoftables

\newpage\thispagestyle{plain}
\section*{List of acronyms}

\medskip
\renewcommand{\arraystretch}{2}
\begin{tabular}{l@{\hspace{1em}}l}
AWS&Amazon Web Services\\
BNC&British National Corpus\\
CBOW&Continuous bag-of-words\\
LSTM&Long Short-Term Memory\\
NLTK&Natural Language Toolkit (Python module)\\
VSM&Vector Space Model\\
\end{tabular}

\newpage\thispagestyle{plain}
\addcontentsline{toc}{chapter}{Abstract}
\section*{Abstract}
Verbal metonymy has received relatively scarce attention in the field of computational linguistics despite the fact that a model to accurately paraphrase metonymy has applications both in academia and the technology sector. The method described in this paper makes use of data from the British National Corpus in order to create word vectors, find instances of verbal metonymy and generate potential paraphrases. Two different ways of creating word vectors are evaluated in this study: Continuous bag of words and Skip-grams. Skip-grams are found to outperform the Continuous bag of words approach. Furthermore, the Skip-gram model is found to operate with better-than-chance accuracy and there is a strong positive relationship (phi coefficient = 0.61) between the model's classification and human judgement of the ranked paraphrases. This study lends credence to the viability of modelling verbal metonymy through computational methods based on distributional semantics.

\newpage
\pagenumbering{arabic}

\chapter{Introduction}

The central question that guides this study, namely ‘to what extent does language have a distributional structure?’, can be thought of as an exploration of what we can discover about the meaning of a word from the language that surrounds it. There is much to be learned about a word from its ‘neighbourhood’. At first this may seem like an obvious proposition, for when reading a text we store in our memory what came before each word. We catch ourselves anticipating subsequent words and thinking of the ways in which elements of language pattern with each other. We notice that not all words are distributed with equal frequency in everyday speech and that certain words tend to cooccur, as happens with idioms. Zellig Harris notes that “All elements in a language can be grouped into classes whose relative occurrence can be stated exactly.” (1954: 146). Harris goes on to say that investigating the occurrence of members of one class relative to those of another class would require the use of statistical analyses informed by an extensive corpus of data. The present study makes use of the British National Corpus (BNC) to create a computational model which paraphrases a particular linguistic phenomenon: that of verbal metonymy.\hfill\break\vspace{5mm}

Metonymy – a type of figurative language – is defined by Shutova et al. as “the use of a word or phrase to stand for a related concept that is not explicitly mentioned” (2013: 11). Verbal metonymy extends this idea, and refers to a use of language where noun phrases are interpreted as events rather than as the objects they usually refer to. An example of this would be the sentence ‘She enjoyed the book’, where an event-selecting verb is combined with an entitydenoting noun. This type clash does not present a problem for native speakers, who will readily accept the sentence. This being said, there is both theoretical (Pustejovsky 1991) and experimental (Lapata et al. 2003) work that investigates how the interpretation of logical metonymy can be influenced by contextual factors, namely the verb’s subject. This study pulls on the thread of these findings about context with the intent of successfully finding the meaning of verbal metonymy by using distributional semantics. Phrases that make use of logical metonymy have been found to present a recurring problem for a compositional parsing of meaning (Bouillon et al. 1992; Pustejovsky 1995) – as such, it is interesting to investigate exactly how it may be possible to paraphrase metonymy. This study understands the main task of a computational understanding of metonymy to be the recovery of the covert event that is not realised in the sentence at a surface level. Taking the aforementioned example of ‘She enjoyed the book’, native speakers of English are likely to assume the covert event to be ‘read’, as in ‘She enjoyed (reading) the book’. This intuition concerning the covert action being carried out also depends on the subject performing the action: chefs are prone to ‘enjoy (cooking) the meal’, whereas most other people are likely to ‘enjoy (eating) the meal’.\hfill\break\vspace{5mm}

For native speakers of English the aforementioned recovery of covert events is a trivial task. This is not so for computers. Despite being increasingly more powerful, faster and cheaper, computers are ill-equipped to handle the subtleties of language. It would seem that there is no such thing as Moore’s Law for making sense of natural language (Moore 1965). That is not to say that the entire domain of language is out of bounds for computers. Quite the contrary – in fact, any computation beyond the punch card is made possibly in part thanks to the fact that there are aspects of language processing at which silicon excels. Abstract syntax tree parsers are a core components in many modern programming languages, for instance, and this deftness carries over when it comes to parsing the syntax of natural language. The rules that govern syntactic derivation are easy for a computer to understand relative to some of the subtleties and ambiguities that semantics has to offer. The reasons why tackling the latter by computational means is worthwhile are twofold. First, studying and expressing ambiguous natural phenomena by using the rigid language of computers is an interesting challenge where there currently exists a gap in the knowledge. Second, it is a stepping stone in the journey towards a more perfect interface between digital repositories of knowledge and the humans who access them by means of natural language queries. To this day, one of the most effective and efficient means of translating native speaker intuitions into code is the Vector Space Model (VSM). Introduced by Salton et al. in 1975, the VSM is a way of representing “the relative importance of the terms in a document” (Manning, Raghavan \& Schütze 2008: 110). Words are assigned vectors based on other words that tend to surround them. Much like how the tradition of gematria assigns numerical values to words in biblical texts – drawing connections between words with equal values – computational linguistics may be thought of as performing a similar task. Once the vector for each word is computed for a corpus, it is possible to carry out vector algebra with these representations of meaning. To quote an example from Mikolov et al., equations such as “Paris - France + Italy = Rome” may be defined thanks to underlying word embeddings (2013b: 9). This word arithmetic has seen wide adoption in the technology sector and academia, with applications ranging from machine translation (Wolf et al. 2014) to visual representations of knowledge (Kottur et al. 2016).\hfill\break\vspace{5mm}

An additional motivating factor that has pushed me to pursue this research is to update some of the practices documented in the existing literature. Some of the earliest work on paraphrasing metonymy was performed two decades ago (Verspoor 1997; Utiyama et al. 2000). Three key technological advances have taken place since which make revisiting computational paraphrasing of verbal metonymy a pressing matter. The first is the advent of cloud computing and the natural progression of computers towards becoming more powerful as time goes on. Second, the release of cutting-edge, more robust dependency parsers such as the new Stanford parser (Manning et al. 2016; Manning \& Schuster 2016). Lastly, the revamped edition of the British National Corpus was released in 2007 in a more accessible XML format (cf. Lapata \& Lascarides’ 2003 paper on metonymy which used an earlier version of the BNC). These factors contribute towards making my study a worthy update to the question of how to best paraphrase verbal metonymy. Moreover, beyond the academic purpose of my research question, this dissertation also aims to document best practices for carrying out Natural Language Processing research using cloud computing techniques.\hfill\break\vspace{5mm}

The main source of data for this study, the British National Corpus (BNC), is a structured set of texts that represent British English at the end of the twentieth century. It features both spoken and written language and totals one hundred million words. The BNC is distributed by the University of Oxford, who also makes available the ‘BNC Baby’, a sample 4\% the size of the full BNC. Both datasets are used in this paper. The distribution of the data in these proportions makes it ideal for training and testing an algorithm. A portion of the larger, full BNC is used to create a vocabulary of vector representations for the top ten thousand most common words. The BNC Baby is then used to test the model and generate the data on verbal metonymy presented in this paper. The creation of word vectors follows in the footsteps of the pioneering work by Mikolov et al., who in 2013 published the models known collectively as ‘word2vec’. More specifically, word2vec presents two alternative ways of generating word embeddings (vectors). The first is Continuous bag-of-words (CBOW), and the second is known as Skip-gram. Before I delve deeper into what either of these do or how they function, it is important to highlight the fact that neither of these qualify as ‘deep learning’. Both algorithms are shallow and reject complexity in favour of efficiency. This does not mean that they are ineffective – quite the contrary. Mikolov et al. (2013b) recommend using Skip-gram as opposed to CBOW. This preference is justified in the context of their study, which concentrates on a phrase analogy task. However, since my research is of a slightly different nature I test both approaches and report on their overall accuracy and the suitability of each for paraphrasing verbal metonymy. Once word vectors have been created from the BNC, the algorithm must look through the BNC Baby for instances of verbal metonymy. Once the set of source sentences which are to be paraphrased has been generated, the next step is to search the BNC Baby again. This time the aim is to find paraphrase candidates whose meaning may approximate that of the original sentence. Once these candidate phrases are validated by a dependency parser, their suitability as paraphrases must be evaluated and each is assigned a confidence score. This scoring is carried out by measuring the cosine similarity between the source metonymy and each of the candidate paraphrases, using data from the pre-computed word vectors. These candidates are then ranked and those with confidence scores above the 0.5 threshold (indicating that the score is better than random chance) are selected as correct paraphrases. The algorithm’s performance is assessed by evaluating it as though it were a binary classifier – one where the labels assigned to the data are either ‘valid paraphrase’ or ‘invalid paraphrase’. The algorithm’s accuracy is calculated by computing its phi coefficient and the performance of CBOW versus Skip-gram is assessed using a precision-recall graph.\hfill\break

\newpage
\chapter{Literature review}

This chapter presents the background reading that helped to define my research question and critically analyses the methods detailed in previous studies of verbal metonymy. Additionally, it is an account of how reading these papers has informed my decisions in terms of the technologies that I have chosen to use in my experiment. The first section provides an overview of existing studies of verbal metonymy in the literature – including theoretical, psycholinguistic and early computational approaches. I then explain how word embeddings are constructed and more generally how to carry out an empirical study of language using the vector space model. Lastly, I assess the performance of a number of algorithms and report how I integrate what I learn from their failures and successes in my study.\hfill\break\vspace{5mm}

\section{Previous accounts of verbal metonymy}

The aim of the study is to develop a model which paraphrases verbal metonymy. Consider the
following sentences:\hfill
\begin{enumerate}[label={(\arabic*)}]
  \item The cook finished eating the meal.
  \item The cook finished the meal.
\end{enumerate}

In sentence (1) the aspectual verb ‘finish’ combines with a verb phrase meaning the event of eating a meal, thus (1) refers to the termination of an event. Contrast this with (2), where ‘finish’ instead combines with a noun phrase referring to a specific meal. The resulting sentence concerns the termination of an unspecified event involving ‘the meal’. Interestingly, the structure of (2), \break [NP [V [ NP ]]], does not include an event whose termination the sentence could be referring to. Arguments that could pair with the aspectual verb ‘finish’ are restricted to those with temporal or eventive meanings. This restriction is not directly satisfied by ‘the meal’, yet human judges are able to make sense of (2). Katsika et al. suggest that the fact that sentences like (2) make sense despite this conflict means that “a temporal/eventive argument is supplied [to aspectual verbs] at some point during the interpretation of the sentence” (2012: 59). Jackendoff describes logical metonymy as an instance of “enriched composition” (1997: 49), and Utt et al. (2013) succinctly define it as consisting of an event-selecting verb combining with an entity-denoting noun. Making sense of sentences like (2) entails the recovery of a covert event (e.g. eating, making, cooking).\hfill\break\vspace{5mm}

My interest in focusing on verbs stems partly from the fact that other aspects of language have received more attention in past computational studies of semantics. Existing computational accounts of metonymy in the literature explore other instances of metonymy, such as those which use toponyms or proper names in general (Markert and Nissim 2006). Psycholinguistic studies conducted on the interpretation of metonymic language include McElree et al. (2001) and Traxler et al. (2002). The latter tested combinations of metonymic and non-metonymic verbs with both entity- and event-denoting nouns (e.g. The cook [finished / saw]V [the meal / the fight]NP). The study found that sentences featuring a metonymic verb and an entity-denoting object (‘The cook finished the meal’ – the coercion combination) involved higher processing costs. The abundance of psycholinguistic studies of verbal metonymy compared to the relative scarcity of papers from a computational or distributional perspective encouraged me to pursue my research question. The frequency with which metonymy happens in natural language and the ease with which humans can interpret it through context and our knowledge of the world also contribute towards making metonymy an interesting phenomenon to model computationally. Despite general metonymy not generally relying on type clashes as much as verbal metonymy does, there naturally exists a relation between the two. Some of the earliest attempts at generating a computational understanding of general metonymy include Lakoff \& Johnson’s 1980 paper and Verspoor’s 1997 study, which searched for possible metonymies computationally yet carried out a paraphrasing task manually. Verspoor’s work is also relevant here since she used a previous version of the British National Corpus. One of the earliest attempts at fully automating the process is Utiyama, Masaki \& Isahara’s 2000 paper on Japanese metonymy. Shutova et al.’s 2012 paper on using techniques from distributional semantics to compute likely candidates for the meanings of metaphors has been a major influence in getting me to think about possible obstacles and improvements in regards to the ranking algorithm. A 2003 paper by Lapata \& Lascarides has been a guiding influence in the creation of my model. For instance, their finding that it is possible to “discover interpretations for metonymic constructions without presupposing the existence of qualia-structures” has led to my model consisting of a statistical learner algorithm and a shallow syntactic parser as opposed to a more contrived solution (2003: 41). When considering the question of which verbs to target in order to search for instances of verbal metonymy in the BNC, Utt et al. (2013) have provided an invaluable starting point. Utt et al. ask: “What is a metonymic verb?” and “Are all metonymic verbs alike?” (2013: 31). They develop empirical answers to these questions by introducing a measure of ‘eventhood’ which captures the extent to which “verbs expect objects that are events rather than entities” (2013: 31). Utt et al. provide both a useful list of metonymic verbs as well as one of non-metonymic verbs. The list builds upon the datasets provided by two previous psycholinguistic studies: Traxler et al. (2002) and Katsika et al. (2012). The existence of this empirical list is useful since it allows me to bypass the ongoing debate regarding whether individual verbs lend themselves to metonymy. This debate has been approached both by theorists (Pustejovsky 1991), psycholinguists (McElree et al. 2001) and computational linguists (Lapata, Keller \& Scheepers 2003. I return to these studies and their relevance in helping me pick relevant verbs in Chapter 3.\hfill\break\vspace{5mm}

\section{Foundations of computational linguistics}

Though it has been echoed many times when introducing the subject of distributional semantics, J.R. Firth’s pithy quip that “You shall know a word by the company it keeps” (1957: 11) remains the best way to describe the field in the fewest number of words. The core idea that meaning must be analysed with context and collocations in mind was put forward by Firth as early as 1935, when he stated that “no study of meaning apart from context can be taken seriously” (1935: 37). The distributional hypothesis implies that it is possible to identify words with similar meanings by looking at items which have similar row vectors when a word-context matrix is constructed. Before proceeding, allow me to illustrate what word vectors are and clarify exactly how they are created. A word vector (a term used interchangeably with ‘word embedding’) is an array of numbers which encodes the context in which a word is typically found in a corpus. For instance, consider the proverb ‘What is good for the goose is good for the gander’. This sentence can be represented as a word-context matrix as shown in Table 2.1. The columns represent each word present in the corpus (Table 2.1 assumes there are no other words in the English language besides those in the proverb). The columns are ordered alphabetically from left to right. The rows represent the words we want to generate vectors for – this usually means each word in the corpus gets its own row, but for illustrative purposes Table 2.1 only generates vectors for ‘good’ and ‘goose’. The numbers at the intersection of two words are generated by calculating \textit{count($w_{i}$ | $w_{i+1}$)}, where $w_{i}$ is the word we want to generate a vector for and $w_{i+1}$ is the word immediately after it. The word ‘for’ occurs twice after ‘good’, which means that the vector for ‘good’ is [2, 0, 0, 0, 0, 0, 0]. Similarly, the vector for ‘goose’ is [0, 0, 0, 0, 1, 0, 0]. Vectors such as the latter, where the only values are one or zero are known as ‘one-hot’ arrays, and I return to them in section 3.1.\hfill\break\vspace{5mm}

\renewcommand{\arraystretch}{0.8}
\begin{table}
\centering
\begin{tabular}{c|ccccccc}
 &for&gander&good&goose&is&the&what\\\cline{1-8}
good&2&0&0&0&0&0&0\\
goose&0&0&0&0&1&0&0\\
\end{tabular}
\captionsetup{width=0.9\linewidth}
\caption[An example of a word-context matrix]{An example of a word-context matrix. This matrix uses the
proverb ‘what is good for the goose is good for the gander’ as a corpus.}
\end{table}

If we were to use the entirety of the BNC as the corpus instead of a single idiom, the vector for good could look something like this: [2, 1, 4, 2, 6, (…)]. When iterating over large datasets the number of rows for which the value is zero is substantial (as can be seen even in the toy example in Table 2.1). To overcome the inefficiency of having arrays full of zeroes and infrequent pieces of actual data, word embeddings are usually stored in what are known as sparse vectors. This means that only columns with non-zero values are stored. Such a vector can have hundreds of rows (also known as dimensions), as is the case of the Google News dataset, which contains vectors for three million words, each vector formed by three hundred rows (Mikolov et al. 2013b: 6). Mikolov et al. reduced the computational complexity of vector generation and released a set of remarkable algorithms when they open sourced their approach to Continuous bag-of-words and Skip-gram under the word2vec tool. They were able to do so by standing on the shoulders of giants, albeit ones who have since received less recognition. Bengio et al. published a paper on probabilistic language models which provided one of the earliest algorithms for generating and interpreting “distributed representations of words” (2003: 7). One of the pioneering outcomes of this paper was defeating the so-called ‘curse of dimensionality’, which Roweis \& Saul had previously attempted to solve (2000). The curse makes reference to the fact that the sequences of words evaluated when implementing an algorithm are likely to differ from the sequences seen during training. Bengio et al.’s use of vector representations trumped prior solutions based on n-gram concatenation both in efficiency and in overcoming the hurdle of the ‘curse’. Another milestone in the path towards word2vec was Franks, Myers \& Podowski’s patent “System and method for generating a relationship network”, published during their time at the Lawrence Berkeley National Lab in 2005 (U.S. Patent 7,987,191). This method is exhaustive and more intricate than word2vec, but ultimately this complexity does not translate into gains in accuracy.\hfill\break\vspace{5mm}

\section{The Vector Space Model}

As mentioned earlier, the techniques introduced by word2vec are not ‘deep learning’ as such. Both the CBOW algorithm and the Skip-gram approach are shallow models which favour efficiency over intricacy. The choice between deep and shallow learning was a consideration made early in the planning stages of this study. Reading Jason Brownlee’s 2014 article on deep learning, in which he speaks of “the seductive trap of black-box machine learning”, was an early indication that a shallow model may be more suitable for my study (Brownlee 2014: 1). Brownlee highlights issues with neural networks, namely that they are by definition opaque processes. Jeff Clune succinctly summarises the issue by saying: “even though we make these networks, we are no closer to understanding them than we are a human brain” (Castelvecchi 2016: 22). Despite Le \& Zuidema’s (2015) recent success in modelling distributional semantics using Long Short Term Memory (a type of recursive neural network), my mind was set against using deep learning in this study for two reasons. First, it would be excessively complicated for the scope of the research being undertaken, and second, the ‘black-box’ nature of neural networks would complicate writing about the inner workings of my algorithm. Having decided between deep and shallow learning and opting for the latter, I faced another choice before creating my model. I had to decide between the two most widely adopted vector space representations: word2vec (Mikolov et al. 2013b) and GloVe (Global Vectors for Word Representation; Pennington, Socher \& Manning 2014). Python implementations of both are available as open source, through Kula’s (2014) glove-python module for GloVe and Rehurek \& Sojka’s (2010) gensim module for word2vec. The documentation for the gensim module characterises the difference between the two technologies by saying that GloVe requires more memory whereas word2vec takes longer to train (Rehurek 2014). Since memory is expensive and time was not a pressing concern, I chose to use the word2vec algorithms (CBOW and Skip-gram) as implemented by the gensim Python module. This decision was supported by Yoav Goldberg’s (2014) case study of the GloVe model. Goldberg disproves Pennington, Socher and Manning’s (2014) claims that GloVe outperforms word2vec by a wide margin and does this by testing both on the same corpus, which the authors of the GloVe paper had neglected to do.\hfill\break

As mentioned earlier, Mikolov et al. (2013b) recommend using Skip-gram as opposed to CBOW in the paper that introduced word2vec. Their preference for Skip-gram is justified by the impressive accuracy improvements they report over earlier work such as Turian et al.’s 2010 paper on word embeddings. This preference is further substantiated by Goldberg \& Levy’s 2014 study on the negative-sampling word embedding algorithm used by the Skip-gram approach. However, I must draw attention to the fact that Mikolov et al.’s original claims are based mainly on a word and phrase analogy task (extending the work first reported in Mikolov et al. 2013a). Since their original findings, impressive as they are, seem to be limited to this linguistic context I cannot presuppose that the Skip-gram approach is necessarily best for all other cases. As such, part of my study is also devoted to testing whether the CBOW or Skip-gram approach is the most suitable for the task of generating word embeddings which successfully paraphrase verbal metonymy. Success in this task is defined as the algorithm which returns the highest proportion of accurate paraphrases. Since this project is being undertaken over a timespan of months, speed is a secondary concern.\hfill\break\vspace{5mm}

Vectors for each word are created by observing the patterns in which a particular word tends to appear. For instance, when generating the vector for the name of a country it is quite likely that the structure ‘citizens of X marched on the streets…’ is present many times in the corpus, where X can be a number of countries. The important factor here is that it does not matter whether ‘France’, ‘Italy’ or ‘Nicaragua’ stand in the place of ‘X’. Rather, what matters is that the algorithm into which the resulting vectors are fed into learns the relationship between each of these words and eventually recognises that they are all instances of the same kind of entity (despite not necessarily knowing that the label that speakers of English assign to these words is ‘country’). An intuitive way of visualising the way in which the algorithm sees these representations of meaning is shown in Figure 2.1. Four vectors are shown in Figure 2.1, which is based on Mikolov et al.’s graphical representation of the Vector Space Model (2013b: 4) and uses data from my model trained on the BNC. Each connects the semantic representation of a country to its corresponding capital city in the vector space. What is of interest is the proximity of each label to each other and the angle at which the connecting vectors (grey dashed lines) are drawn. By observing the proximity of labels, we can intuitively tell that Spain, Italy and France are closer neighbours to each other than Nicaragua is to either of the three European countries. However, if Figure 2.1 were to show the entirety of the BNC, we would see that indeed Nicaragua is closer to Italy, for instance, than it is to ‘herring’. The grey lines, the vectors connecting word embeddings, are of relevance when seeking to evaluate the similarity between two entities in the vector space. By computing the cosine similarity between the two lines, normalised similarity scores between the semantics of each word can be obtained (the method and consequences of doing so are explored more in depth in section 3.3). More intuitively, by looking at Figure 2.1 it is evident that the lines have similar angles and bearings, and as such must bear some similarity in the semantic relations they encode.\hfill\break\vspace{5mm}

\begin{figure}[]
\centering
\includegraphics[width=0.5\textwidth]{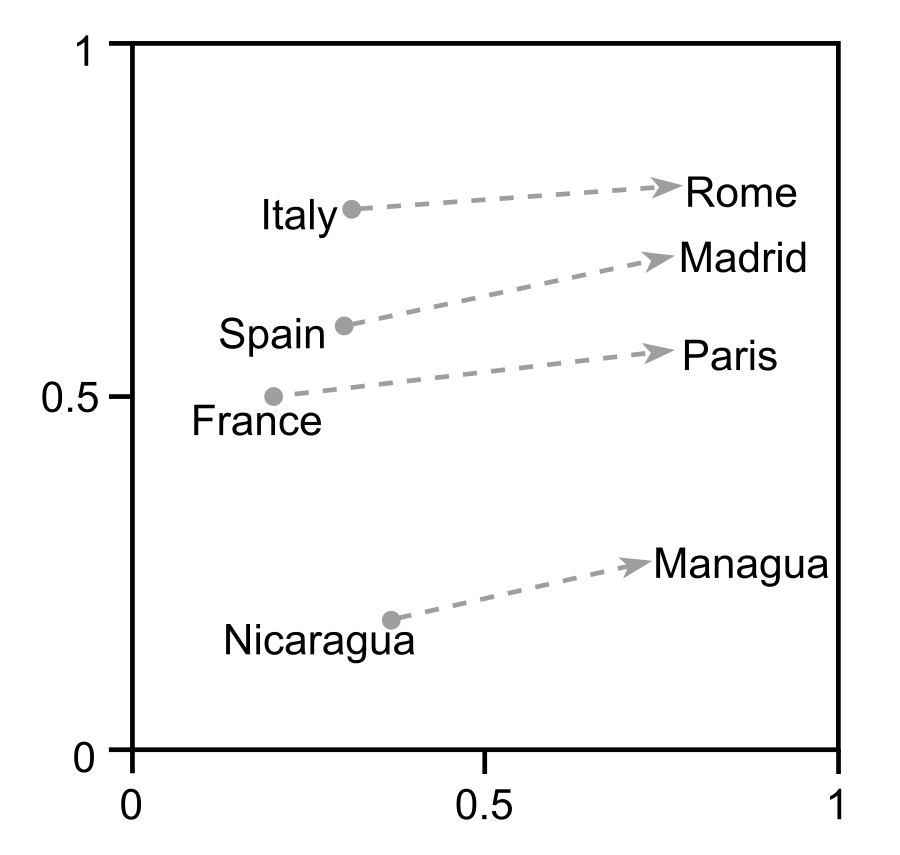}
\captionsetup{width=0.9\linewidth}
\caption[A vector space model of country-capital city relations]{A vector space model of country-capital city relations. This model was generated using the Skip-gram approach trained on data from the British National Corpus.}
\end{figure}

Besides calculating similarity scores it is also possible to carry out vector algebra with these representations of meaning. Equations such as the aforementioned “Paris - France + Italy = Rome” (Mikolov et al. 2013b: 9) are interesting, but this paper makes recurrent use of cosine similarity instead. Mikolov et al.’s stringent accuracy metrics (they accept only exact matches) mean that the overall accuracy of this semantic algebra stands at 60\% in their original implementation (2013b: 10). However, more recent studies have refined such algorithms and accept ‘closest-neighbour’ answers rather than limiting themselves to exact matches (Levy, Goldberg \& Ramat-Gan: 2014; Gagliano et al. 2016). The present study rejects the hindering stringency of Mikolov et al. and instead uses a ‘closest-neighbour’ evaluation.\hfill\break\vspace{5mm}

\newpage
\chapter{Method}

In this chapter I provide an account of how I determined whether the distributional hypothesis can be used to paraphrase verbal metonymy by recovering covert events. The starting point for my research is the British National Corpus, a dataset that is present throughout this study. It is used to first train the model by generating a vocabulary of word embeddings and then employed in searching for paraphrase candidates and evaluating them. Section 3.1 explores what the BNC is and how it is structured, as well as explaining how I implemented the CBOW and Skip-gram algorithms in order to create word vectors. The next section details the search for instances of metonymy in the corpus which are then used as the target phrases. In this context ‘target’ refers to phrases such as “I think you should begin the next chapter now” (Appendix I: begin-4), which are the sentences containing verbal metonymy that I wish to paraphrase. I refer to the phrases that could potentially paraphrase the target as ‘candidates’ (for the above target ‘begin the chapter’ candidates include: ‘read the chapter, write the chapter, etc.’). Finally, section 3.3 describes the procedure through which I found these candidates, validated them using dependency parsing and generated a confidence score for each with which to rank them. The method described in this chapter was implemented using servers hosted by the cloud computing provider Amazon Web Services (AWS). The specifics of the software versions and hardware specifications can be found in Appendix II. Details of server configuration tailored towards replication of this study are made available in Appendix III. Additionally, all the code used in my research is made available under an open source license and can be located using the Digital Object Identifier \texttt{doi:10.5281/zenodo.569505}.\hfill\break\vspace{5mm}

\section{The BNC and word2vec}

The British National Corpus is a dataset of British English during the second half of the twentieth century. The corpus consists of one hundred million words split across a number of genres, with 90\% of the data comprising the written part of the BNC, while the rest is composed of transcriptions of spoken English (Burnard 2007). The corpus also includes data from an automatic part-of-speech tagger. Text samples do not exceed forty-five thousand words each and were collected from a variety of mediums writing in a number of genres. The first version of the BNC was released in 1994, with subsequent revisions appearing in 2001 and 2007. This study makes use of the 2007 BNC in XML format. The BNC has been used in studies covering a variety of disciplines, including syntax (Rayson et al. 2001), sociolinguistics (Xiao \& Tao 2007) and computational linguistics (Verspoor 1997; Lapata \& Lascarides 2003). Besides being a reputable corpus which has been implemented in a number of studies, a major reason for choosing the BNC is of a more practical nature. The Natural Language Toolkit (a module extending the Python programming language) includes an interface for efficiently iterating over the BNC. Due to my intention to use Python as the main language in this project and my previous experience with the NLTK module, the choice of the BNC as a source of data was an obvious one. I still had to write code of my own with which to parse the corpus, find cooccurrences and create vectors, but the use of NLTK sped up the process considerably. A four-million-word sample (known as the BNC Baby) is available alongside the full BNC. This sample contains the same proportion of spoken and written texts, and the distribution of texts by genre and domain remains the same as in the full corpus. This proportion of data between the BNC Baby and the full corpus makes it ideally split for creating a training set and a test set. In this study, a ten-million-word fragment of the full BNC (different from the words in the BNC Baby) is used to generate the word embeddings on which the model is subsequently trained. The purpose of the training data is to discover relationships between words in the corpus, and as such would be bad practice to test an algorithm on the same data used to train it. Neglecting to do this usually leads to overfitting the data, which means that the model learns too much about the random variation it should not be interested in rather than focusing on the actual relationships (Wei \& Dunbrack 2013).\hfill\break\vspace{5mm}

The first step in this study is the generation of two vocabularies of word embeddings for the data found in the BNC: one generated using word2vec’s Continuous bag-of-words, the other using Skip-grams. First, I analyse the CBOW approach. Consider the following sentence: ‘Those who cannot remember the past are condemned to compute it’ (Pinker 1999: 164). The first step CBOW takes in order to generate word vectors from this sentence is to ‘read’ through it one word at a time, as through a sliding window which includes a focus word together with the four previous words and the next four words. This would mean that for the focus word ‘past’, its context window is formed by ‘who cannot remember the condemned to compute it’. A window size of four context words was chosen on account of Shutova et al.’s (2012) experimental success with a smaller window size than the one used by Erk and Padó (2008). Additionally, Mikolov et al.’s original paper also uses four words and warns that window size is one of the most important factors that affect the performance of a model (2013b: 8). The context words are encoded in a ‘one-hot’ array, as seen previously in the example for ‘goose’ in Table 2.1. The number of dimensions was set to one hundred columns. Additionally, a weight matrix is constructed, which is a representation of the frequency of each word in the corpus. This weight matrix has V rows, where V is the size of the vocabulary and D columns, where D is the size of the context window, also referred to as the number of dimensions (in my implementation, eight dimensions). The weight matrix does not represent a one-to-one relation between values in the rows and the word the row represents. Rather, the representation of the word is scattered amongst all the columns in the array. In the example of the quote by Pinker, each one-hot array would have eleven dimensions, with only one of its columns set to one, the rest to zero. Once the arrays and weight matrix have been created, the\break algorithm trains the model with the aim of maximising\linebreak P($w_{f}$ | $w_{f-4}$ … $w_{f+4}$). That is, maximising the probability of observing the focus word, $w_{f}$, given the eight context words surrounding it. In our example the objective of training is to maximise the probability of ‘past’ given the eight words in the context window as an input. Table 3.1 shows a one-hot array with eleven dimensions (vocabulary size) being multiplied by the weight matrix for the corpus (this matrix is truncated and only shows the first three of eleven rows, but does show the full number of dimensions: eight, corresponding to the number of context words). CBOW computes the final vector for each word it encounters by performing this operation many times. Finally, a normalized exponential function (also known as the softmax function) is used to produce a categorical distribution: a probability distribution over D dimensions (Mikolov et al: 2013a; Morin \& Bengio 2005).\hfill\break\vspace{5mm}

The Skip-gram model, on the other hand, takes the Continuous bag-of-words approach and effectively turns it on its head. Where CBOW uses eight one-hot context word arrays as inputs, Skip-gram uses a single array. This input vector is a one-hot array of size D constructed with the focus word instead (‘past’ in the example above). The same process involving the weight matrix is used, but this time the aim is to output the probability of observing one of the context words. Where CBOW output a single probability distribution, Skip-gram outputs eight different ones. This last step is quite resource intensive, particularly in regards to memory. An efficient and effective solution proposed by Rong is to “limit the number of output vectors that must be updated per training instance” (2014: 10). This is achieved using the softmax function again. Softmax represents all the words in the vocabulary as elements of a binary tree and computes the probability of a random walk from the root to any word in the vocabulary. The further intricacies of this approach are beyond the scope of the present paper. Instead, I direct the reader’s attention to the work of Morin \& Bengio (2005), Mnih \& Hinton (2009) and the aforementioned paper by Rong (2014) which explains in great detail the many parameters of word2vec and the potential optimisations that may be applied to CBOW and Skip-gram. The main advantage of implementing Skip-grams with the improvements suggested by Rong is that there is a boost to speed without a loss of accuracy. Instead of having the Skip-gram algorithm evaluate V output vectors, it only has to process $log_{2}(V)$ arrays instead (Rong 2014: 13). For the Pinker example, this means going from 11 vectors to $log_{2}(11)$ $\approx$ 3.46 vectors – a considerable difference since there are 68\% less arrays to evaluate. Despite the obvious improvements offered by Skip-grams, two separate training sets were created by running CBOW and Skip-grams on a ten-million-word fragment of the British National Corpus. These sets have a vocabulary size of the ten thousand most frequent words in the BNC and the arrays have one hundred dimensions (columns). As a benchmark, this may be compared to the size of the Google News vector dataset. This dataset has become one of the standards for collections of word embeddings in academia and the open source community. Released by Mikolov et al. (2013b), it has a vocabulary composed of the 1 million most frequent words in Google News articles, with each array comprising three hundred dimensions.\hfill\break\vspace{5mm}

\renewcommand{\arraystretch}{0.8}
\begin{table}
\centering
\begin{tabular}{ccc}
input&weight matrix&hidden layer\\
1 x V&V x D&1 x D\\
\lbrack\ 0 0 1 0 0 0 0 0 0 0 0 \rbrack\ $\bullet$ &$\left[ \begin{array}{cccccccc}
a,&b,&c,&d,&e,&f,&g,&h\\
\color{orange}i,&\color{orange}j,&\color{orange}k,&\color{orange}l,&\color{orange}m,&\color{orange}n,&\color{orange}o,&\color{orange}p\\
q,&r,&s,&t,&u,&v,&w,&x\\
 & & & & \dots & & & \end{array}\right]$&= \lbrack\ \color{orange}i, \color{orange}j, \color{orange}(...) \color{orange}o, \color{orange}p \color{black}\rbrack \\
\end{tabular}
\captionsetup{width=0.9\linewidth}
\caption[One-hot arrays in the CBOW algorithm]{One-hot vectors in the CBOW algorithm. V is the vocabulary size and D is the size of the context window. The values ‘a’ through ‘x’ represent the distribution of weights assigned as a function of each word’s frequency in the vocabulary. (Adapted from Colyer 2016).}\hfill
\end{table}

\section{Searching for metonymy}

Once the two training datasets have been generated, they are not needed again until section 3.3, where they are necessary in order to evaluate the paraphrases generated by the model. The next step is to decide the kind of metonymy that the model should aim to paraphrase and look for examples in the test data. The test data is composed of the entirety of the four-million-word BNC Baby. In order to keep the set of sentences to paraphrase and the number of candidates returned by the model manageable, this experiment only considers instances of metonymy which employ one of three verbs. These three verbs are ‘begin’, ‘enjoy’ and ‘finish’. There are two reasons why these three verbs have been selected. The first is so that verbs from both categories defined by Katsika et al.’s 2012 psycholinguistic study on complement coercion are present. ‘Begin’ and ‘finish’ are metonymic aspectual verbs while ‘finish’ is a metonymic psychological verb (Katsika et al. 2012: 61). Secondly, these three verbs are uniformly distributed across the data presented by Utt et al. (2013). Their paper assigns numerical values to a measure of ‘eventhood’ which captures the extent to which these verbs “expect objects that are events rather than entities” (2013: 31). ‘Begin’ receives an eventhood score of 0.91, ‘finish’: 0.66 and ‘enjoy’ scores 0.57 (the upper bound for eventhood was 0.91, the lower 0.54) and are all confirmed to take part in metonymic constructions (Utt et al. 2013: 7).\hfill\break\vspace{5mm}

First, the BNC Baby is scraped for sentences which contain one of the three target verbs. This aims to cut down on the processing costs of any subsequent tasks so that it is not necessary to iterate over irrelevant sections of the corpus. Next, the algorithm looks through these files for instances of Noun Phrases present immediately after or in close proximity following one of the three verbs – these are potential instances of metonymy. This is facilitated by the fact that the BNC features an extensive amount of metadata in the tags for each word. Once the list of all sentences which potentially contain verbal metonymy has been created, sentences are inspected manually to discard false positives where there is no target to paraphrase (naturally, it would be ideal to automate this task, and this is considered in the future directions evaluated in Chapter 6). However, it is crucial that these target sentences are actual instances of metonymy, and as such required human judgement. The task described in section 3.3, the actual generation and ranking of paraphrases, is completely automated. The list of target sentences is scraped for the noun phrases that are typically used in conjunction with the three verbs when they are used metonymically. Once this is done, the entirety of the BNC Baby is searched through for sentences containing the noun phrases observed in the previous step. This approach of concentrating on collocations is supported empirically by the results reported by Verspoor (1997), who found it to be the case that “95.0\% of the logical metonymies for begin and 95.6\% the logical metonymies for finish can resolved on the basis of information provided by the noun the verb selects for” (Lapata \& Lascarides 2003: 41).\hfill\break\vspace{5mm}

\section{Generating paraphrases}

Once the model has observed all the sentences in the BNC Baby that contain noun phrases commonly associated with instances of verbal metonymy involving one of the three verbs, it is time for the last two steps of the algorithm. This section contemplates the generation of paraphrases and the subsequent assignment of a confidence score to each of these candidates so that they may be ranked. The first task, that of generating the paraphrases, iterates through the sentences collected at the end of the last section and validates them with the Stanford parser. For example, suppose that the algorithm aims to paraphrase “He seems to enjoy the job, doesn't he?” (Appendix 1: enjoy-3). It looks through the BNC Baby looking for collocates of the noun phrase “the job” and returns candidate sentences that include constructions such as “get / do / see the job”. Once this is done, each candidate is considered separately and submitted to the updated Stanford dependency parser (Manning et al. 2016). Originally released by Marneffe et al. (2006), the parser provides “both a syntactic and a shallow semantic representation” (Manning \& Schuster 2016: 1). The parser outputs typed dependencies (grammatical relations) between the elements of any string provided as input. In this model, it performs the task of checking that the verb and noun in the candidate paraphrase are in a direct object relationship. The final step in the algorithm is to compute the confidence score for any approved candidates. This is done by computing the cosine similarity between the joint word vector of the target phrase and that of the candidate. The similarity is obtained by dividing the dot product of the two vectors by the product of the two vectors’ magnitudes, or expressed as a formula:\hfill

\begin{gather*}
cos\mathit{\theta}  =
\frac{
\mathit{\boldsymbol{V_{1}}} \cdot \mathit{\boldsymbol{V_{2}}}}
{\left \| V_{1} \right \|
\left \| V_{2} \right \|}
\end{gather*}\vspace{2mm}

A simple explanation for this formula is that the numerator measures the degree to which the two vectors are related, while the denominator serves as a normalization factor that keeps the result under a maximum value of one. Figure 3.1 shows this graphically – the angle $\theta$ between two vectors is measured and the cosine of the angle gives a value which determines how related the two vectors are. The function returns a minimum value of zero corresponding to vectors perpendicular to each other, meaning they are unrelated. Values range up to one, only obtained when comparing identical vectors.\hfill

\begin{figure}[H]
\centering
\includegraphics[width=0.5\textwidth]{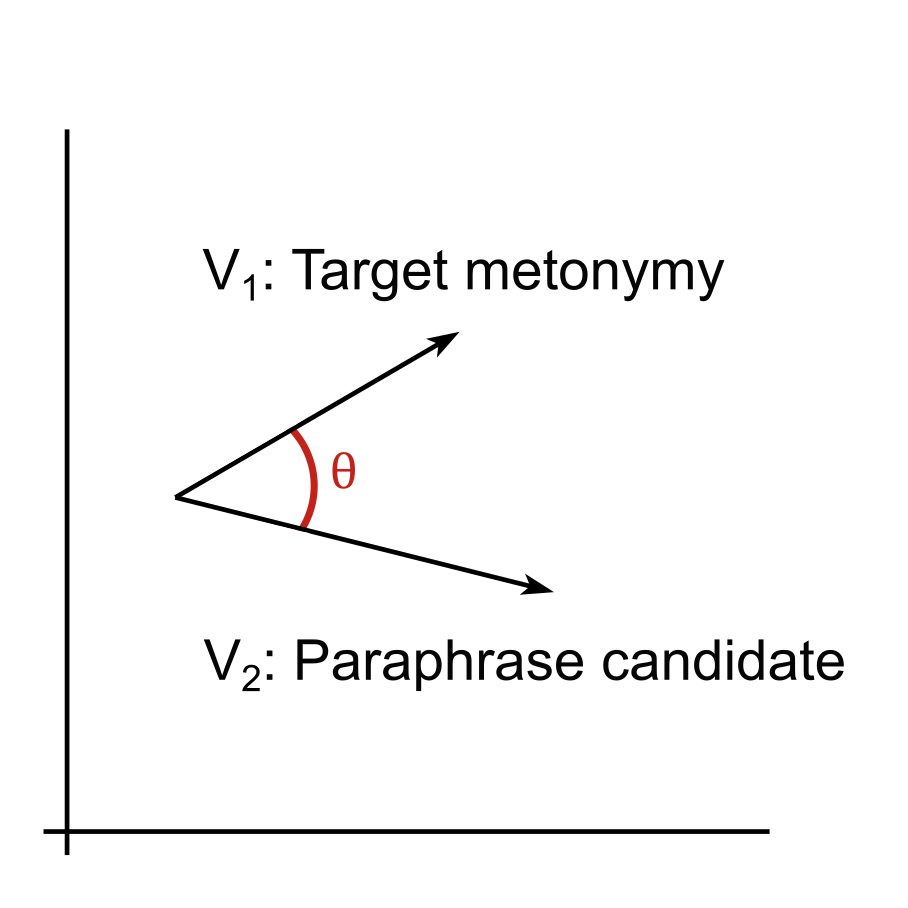}
\captionsetup{width=0.9\linewidth}
\caption[Cosine similarity between word vectors]{Cosine similarity between word vectors. Each sentence is represented by a vector and the cosine of the angle between them yields a similarity score.}
\end{figure}

The model generates similarity scores for each paraphrase, rejecting those that score below 0.2 as completely irrelevant. Sentences with scores above 0.5 are considered viable paraphrases. Two rankings are created: one that measures similarity against the CBOW word embeddings generated earlier and another that does the same but uses vectors produced using the Skip-gram approach. The algorithm has reached the end of its cycle. The next chapter presents the results of this study, highlighting the successes and pitfalls of the model.\hfill

\chapter{Results}

This chapter presents the outcome of the algorithm described above. First, it is important to note that all the data presented henceforth is the result of training the model on vectors generated using the Skip-gram approach to word embeddings. When comparing the two implementations of word2vec, I quickly found that my results agreed with Mikolov et al.’s assessment that Skip-grams provide better accuracy than CBOW (2013a: 10). I later compare CBOW to Skip-grams and empirically evaluate the performance of each approach. In this chapter I first give a general account of the data obtained by running my algorithm on the BNC Baby. This is followed by a more in-depth look at the data by means of analysing individual target sentences and their paraphrases. The data for individual examples I draw attention to is presented in the form of a table for each target sentence. These tables consist of an example of verbal metonymy which is to be paraphrased, followed by a number of candidates which are ranked and for which I list individual confidence scores. Though the system returns a gradient measure of confidence, the best way to evaluate the algorithm's performance is to treat it as performing a binary classification task where the proposed paraphrases are either relevant or they are not. Candidates with scores above 0.5 are marked as viable paraphrases by the algorithm (denoted in the tables by a green background). Those with a score below 0.5 are not considered accurate paraphrases of the target sentence. Occasionally, some scores are preceded by an asterisk and set in bold type. This indicates a false result – a false positive if on a green background, a false negative when on a grey background. Under the target sentence there is a reference to where in the BNC dataset this phrase was located. It follows the format (\textit{category/FILENAME}, sentence: \#) where category is one of the four categories that the BNC Baby is divided into, filename the XML file in which it was found, and \# is its position in the file. Additionally, there are mentions of \textit{verb-\#} notation for each table, referring to where in Appendix I this table may be found. Appendix I features all the ranking tables for the experiment.\hfill\break\vspace{5mm}

The model first scraped 2,621 potential target phrases out of a total of 332,963 sentences in the British National Corpus. These potential target phrases are those which mentioned ‘begin’, ‘enjoy’ or ‘finish’ in any of their forms. Out of these, 41 sentences were selected as instances of verbal metonymy and stored to be subsequently paraphrased. In order to generate paraphrases, 23,029 sentences (out of 332,963) were selected computationally on the basis of featuring matching noun phrases with the aforementioned 41 target sentences. These were vetted by the Stanford dependency parser, which whittled down the final number to 179 candidatesto be scored. To summarise: the final experiment featured 41 instances of metonymy and 179 paraphrases, an average of four paraphrases per original sentence. The candidates were then assigned confidence scores and ranked. The results of the algorithm’s efforts to label candidates as either valid or invalid paraphrases can be split four ways. Each result can be a true positive, a true negative, a false positive or a false negative. True positives and negatives are those where human judgement agrees with the algorithm’s decision to label a paraphrase as valid or invalid respectively. False positives and negatives are those where human judgement disagrees with the model’s output. Figure 4.1 shows an analysis of these results for each of the three verbs. Out of a total of 179 paraphrases 52 were true positives and 94 were true negatives, while false positives and false negatives accounted for 15 and 18 sentences respectively. Out of the 41 sentences, all of the paraphrases suggested for five of these fell below the 0.5 relevance threshold. Conversely, in the case of four target sentences, all of the paraphrases suggested had confidence ratings above 0.5.\hfill

\begin{figure}[H]
\centering
\includegraphics[height=0.5\textwidth]{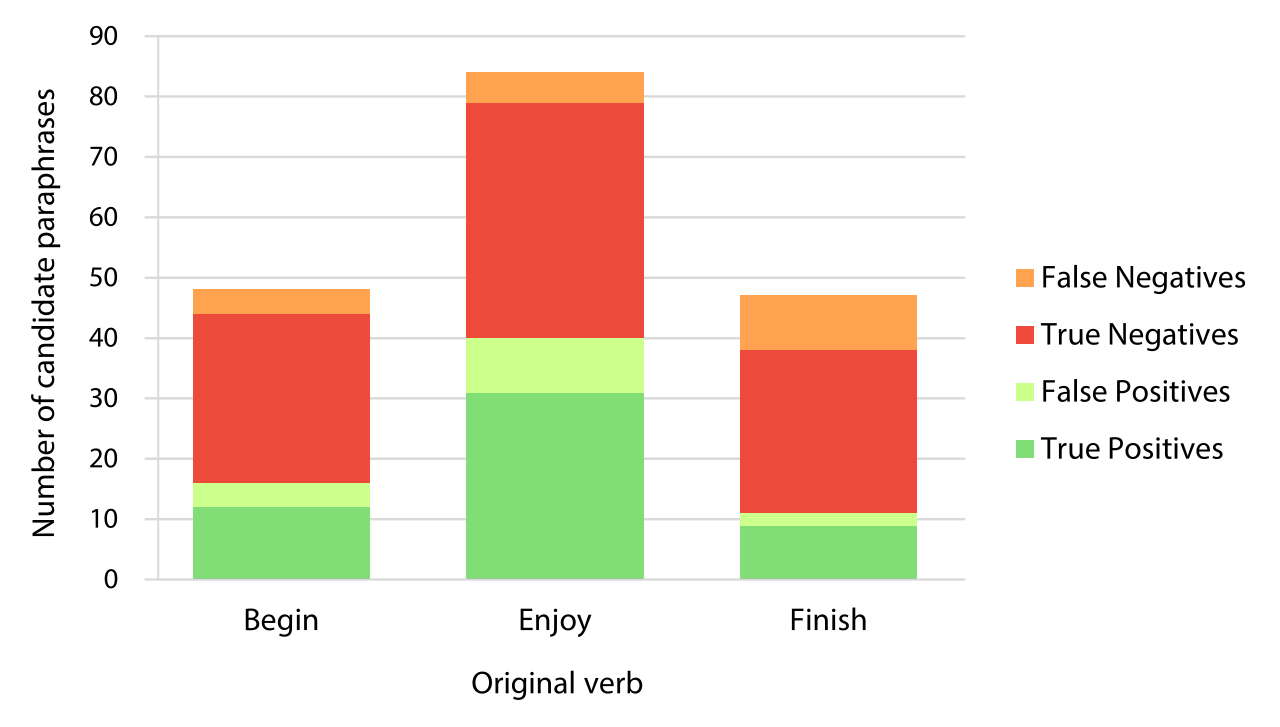}
\captionsetup{width=0.9\linewidth}
\caption[Analysis of paraphrases for the three target verbs]{Analysis of paraphrases for the three target verbs. A total of 179 paraphrases were scored and ranked by the model.}
\end{figure}\hfill

Table 4.1 shows two such outliers: a ranking with perfect scores and another which does not generate a single viable paraphrase. In the table on the right there is in fact a paraphrase that is evaluated as correct by human judgement, but which the model failed to score as relevant. This has been labelled a false negative as denoted by the preceding asterisk and bold type.\hfill\break\vspace{5mm}

\renewcommand{\arraystretch}{0.8}
\begin{table}[H]
\centering
\begin{tabular}{ccc|cc|}
\cline{1-2} \cline{4-5}
\multicolumn{2}{|c|}{\begin{tabular}[c]{@{}c@{}}“I've enjoyed the concert but…”\\ (dem/KPU, sentence 975)\end{tabular}}      &  & \multicolumn{2}{c|}{\begin{tabular}[c]{@{}c@{}}“Finish the last packet of cigarettes…”\\ (news/BM4, sentence 1431)\end{tabular}} \\ \cline{1-2} \cline{4-5} 
\multicolumn{1}{|c}{\cellcolor[HTML]{BBDAFF}Paraphrase candidate} & \multicolumn{1}{c|}{\cellcolor[HTML]{BBDAFF}Confidence} &  & \cellcolor[HTML]{BBDAFF}Paraphrase candidate                     & \cellcolor[HTML]{BBDAFF}Confidence                            \\  
\multicolumn{1}{|c}{See the concert.}                             & \multicolumn{1}{c|}{\cellcolor[HTML]{9AFF99}0.68158}    &  & Carry the packet.                                                & \cellcolor[HTML]{EFEFEF}0.44237                               \\ 
\multicolumn{1}{|c}{Listen to the concert.}                       & \multicolumn{1}{c|}{\cellcolor[HTML]{9AFF99}0.58792}    &  & Crumple the packet.                                              & \cellcolor[HTML]{EFEFEF}0.40580                               \\ 
\multicolumn{1}{|c}{Go to the concert.}                           & \multicolumn{1}{c|}{\cellcolor[HTML]{9AFF99}0.55673}    &  & Smoke the packet.                                                & \cellcolor[HTML]{EFEFEF}\textbf{*0.35518}                     \\  \cline{1-2}
                                                                   &                                                         &  & Open the packet.                                                 & \cellcolor[HTML]{EFEFEF}0.36162                               \\ \cline{4-5} 
\end{tabular}\break
\captionsetup{width=0.9\linewidth}
\caption[Two rankings showing perfect and failing outliers]{Two rankings showing perfect and failing outliers. The table on the left can be found in Appendix I as enjoy-7, the one on the right under finish-11.}
\end{table}

The model had difficulty parsing the true meaning of five paraphrase candidates, all of which contained phrasal verbs or idioms. Though it failed at assigning them scores above the 0.2 discard threshold, I have kept these five instances in the dataset, partly as a testament to the algorithm’s ability to recover more complex structures. They are labelled ‘Not in vocabulary’ since the algorithm was not attempting to paraphrase the whole phrase, and usually defaulted to giving the score for the verb. These five failures to understand the complexities of English are “Set out the research” (found in Appendix I: begin-1), “Keep the process going” (begin-2), “Turn his hand to the task” (begin-6), “Keep an eye on the scene” (enjoy-14) and “Wade through the book” (finish-3). This is also seen in paraphrases that were not discarded since they scored higher than 0.2, but nonetheless were not selected since their score was lower than 0.5. One such case is “Toss back the whisky” for the target sentence “She finished the whisky” (Appendix 1: finish-9). Though the meaning of the paraphrase approximates that of the desired covert event (drink), the algorithm evaluated ‘toss’ over the compositional meaning of ‘toss back’. While in this case other paraphrases included ‘drink’ and ‘gulp’, it remains the case that some idioms remain a stumbling block in the way towards a more accurate algorithm. Conversely, there are some paraphrases which suggest that the algorithm is aware on some level of the existence of these idioms. For instance, consider the data in Table 4.2, which shows a ranking of paraphrases for “finish the job”. There is a single paraphrase that is evaluated as correct, and that is “do the job”. The paraphrase ranked second is “get the job”. Though the ranking could be due to the relative frequency of how often ‘getting a job’ and its variants appear in the BNC, another possibility is that by suggesting ‘get’ for this particular paraphrase, it is trying to reach ‘get on with (something)’. Another fact that also supports this theory about the model’s intuitions regarding idioms is the existence of other expressions that use ‘get’ such as ‘get (the job) over and done with’. This idiom actually paraphrases the target meaning of “finish” more closely than ‘get on with the job’ does.\hfill\break\vspace{5mm}

\renewcommand{\arraystretch}{0.8}
\begin{table}[H]
\centering
\begin{tabular}{|cc|}
\hline
\multicolumn{2}{|c|}{\begin{tabular}[c]{@{}c@{}}“...I want to stay on here to finish the job.”\\ (news/CH3, sentence 307)\end{tabular}} \\
\rowcolor[HTML]{BBDAFF} 
Paraphrase candidate                                      & Confidence                                                                  \\ 
Do the job.                                               & \cellcolor[HTML]{9AFF99}0.58553                                             \\
Get (on with?) the job.                                   & \cellcolor[HTML]{EFEFEF}\textbf{*0.48158}                                   \\ 
Find the job.                                             & \cellcolor[HTML]{EFEFEF}0.42892                                             \\ 
Have the job.                                             & \cellcolor[HTML]{EFEFEF}0.37238                                             \\ \hline
\end{tabular}\break
\captionsetup{width=0.9\linewidth}
\caption[Paraphrase ranking for “finish the job”]{ Paraphrase ranking for “finish the job”. The suggestion “get the job” may indicate an awareness of the idioms such as ‘get on with (the job)’ (Appendix I: finish-10).}
\end{table}

In an aim to maximise simplicity, the method through which sentences that potentially contain verbal metonymy are scraped was a naïve one at the start of development. The method was eventually refined, but an early pitfall was the model returning segments such as “…How about you?’ began the top man…”. While this sentence does indeed contain a covert event that could have been recovered (said or spoke), my algorithm’s focus of verb-noun phrase collocations made the inclusion of this sentence problematic since it would try to interpret ‘*begin the top man’ rather than the actual ‘the top man spoke’. Additionally, during an early stage, the algorithm would occasionally attempt to find synonyms for the verb in the target sentence (i.e. ‘begin’, ‘enjoy’ or ‘finish’) instead of finding the meaning of the covert event. Highly ranked paraphrases for phrases such as “enjoy the countryside” would include verbs such as ‘love’, ‘appreciate’ or ‘savour’ as opposed to the desired paraphrase of “[enjoy] visiting / being (in) the countryside”. While steps were taken to reduce this effect, it was not entirely compensated for and as such still appears to a degree in some ranking tables (e.g. Appendix 1: enjoy-19; the aforementioned ‘countryside’ example).\hfill\break\vspace{5mm}

\chapter{Discussion}

The best way to evaluate a binary classifier, such as the model described in this paper, is by means of a precision-recall graph. Such a graph provides an empirical, quantitative account of an algorithm’s performance and is used here to compare Continuous bag of words and Skip-gram. Figure 5.1 shows two curves. The orange line tracks the performance of my algorithm when fed a vocabulary of word embeddings created using CBOW and the British National Corpus. The blue line shows the model’s performance when using word vectors created using the Skip-gram approach. They are plotted in a way that measures precision against recall. Precision is a way of answering the question ‘what percentage of positive predictions were correct?’. Precision is calculated by dividing the number of relevant items retrieved (true positives) by the total number of items retrieved (the sum of true positives and false positives). Recall evaluates the capacity of a system to return all relevant items, and is measured by dividing the number of relevant items retrieved (true positives) by the total number of relevant items present in the dataset (the sum of true positives and false negatives). Evaluating my algorithm involves plotting precision against recall every time a paraphrase is generated. As such, each line in Figure 5.1 is made up of 179 discrete data points. Precision-recall graphs show how precision degrades over time, as more queries are handled until recall reaches one, which means all queries have been processed. In the case of Figure 5.1, recall reaching one means that all the paraphrase candidates that the algorithm can generate have been generated. The first piece of information that can be gleaned from the precision-recall graph is that the model that used Skip-grams had far better performance than the one using the CBOW method. The results of my experiment agree with Mikolov et al.’s evaluation of CBOW being a weaker algorithm (2013b). My research is another piece of evidence pointing at the superiority of Skip-grams. However, I am wary about making any sweeping statements about the applicability of this trend to the entirety of algorithms modelling English since both my study and Mikolov et al.’s each consider only small niches within Natural Language Processing (phrase analogy and verbal metonymy, respectively). As such it would be wise to wait for data from a wider variety of studies that continues to confirm this trend.\hfill\break

\begin{figure}[H]
\centering
\includegraphics[width=1.0\textwidth]{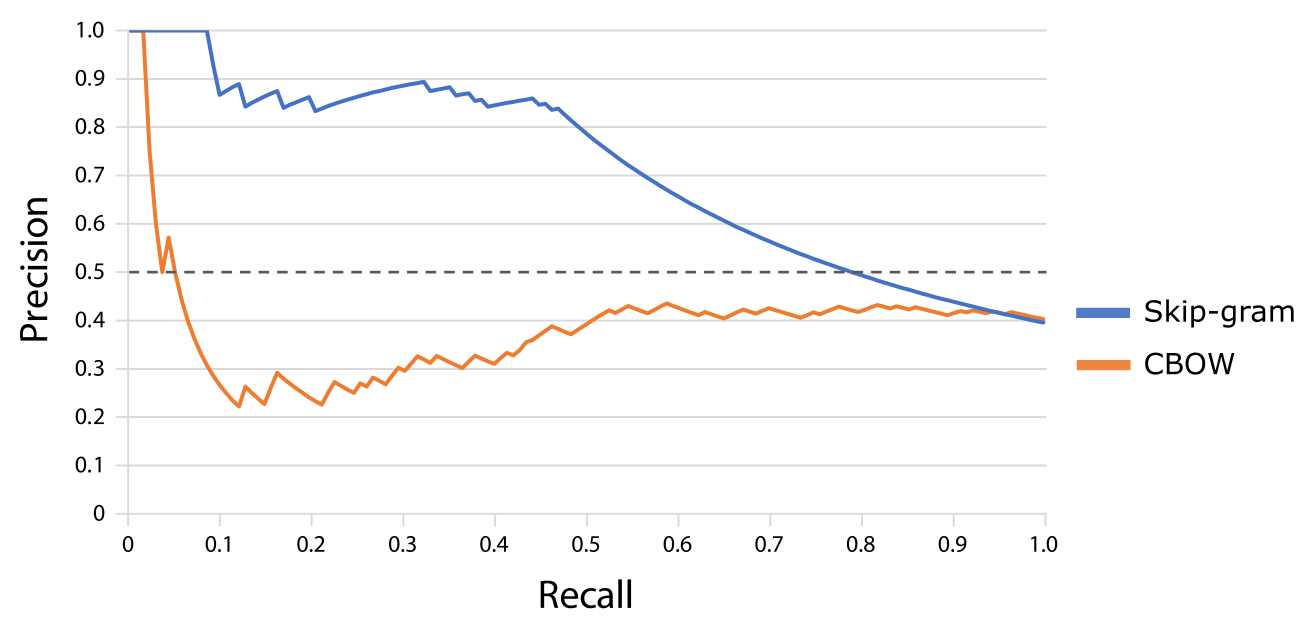}
\captionsetup{width=0.9\linewidth}
\caption[Precision-recall graph comparing CBOW and Skip-gram]{Precision-recall graph comparing CBOW and Skip-gram. Algorithms above the dashed line are performing better than chance.}
\end{figure}\hfill

There is a third line on Figure 5.1, a dashed grey line at y = 0.5 which represents the baseline against which I have measured both algorithms. Any line above the 0.5 threshold is performing at a rate that is better than chance. This adds another condition with which to evaluate CBOW and Skip-gram. While both models eventually end up performing below the threshold, the difference between when each crosses the 0.5 threshold is notable. Both algorithms start by performing at perfect accuracy. After the first 2\% of paraphrases are generated, CBOW begins a swift dive towards the chance threshold. There is a slight recovery before it plummets again by the time it has generated the first 10\% of paraphrases. Skip-gram, on the other hand, maintains perfect accuracy for the first 8\% of the data it outputs. Accuracy then starts to degrade steadily for a few percentage points until it achieves an equilibrium that keeps it hovering between 80\% and 90\% precision. This trend is maintained until Skip-gram has generated over 45\% of the paraphrases it will generate during its execution. After this a steady slump downwards starts, with the precision-recall line intersecting the 0.5 threshold when Skip-gram has generated 79\% of its total paraphrases. Only after 95\% of the total data has been output does Skip-gram’s performance degrade to equal that of CBOW. Another important consideration when assessing which of the two algorithms to use is the execution time for each. As mentioned previously, there are significant differences between how long each take, mainly due to the optimisations introduced by Rong (2014). Each set of word vectors, regardless of being generated by CBOW or Skip-gram, had a vocabulary size of the most common ten thousand words. The running time of CBOW increases geometrically with V, where V is the vocabulary size, whereas Skip-gram’s execution follows log$_{2}(V)$. Since log$_{2}(10$000) $\approx$ 13.3, this means that Skip-gram runs three orders of magnitude faster than CBOW in this experiment. While the outstanding performance of Skip-gram is quite a welcome improvement, I would still caution against implementing it directly in a live system (such as a web application or chatbot). The performance results reported here are acceptable for academia and within the context of a research project. Were this technology to be used in developing a product, however, I would suggest pre-generating the confidence scores for a large corpus of data (as this experiment does with word vectors) rather than generating them synchronously with user input. The use of an updated dependency parser (cf. Lapata \& Lascarides’ use of the Cass parser in 2003) also meant some improvements in filtering relevant instances of metonymy from extraneous data. Despite this, some issues still remain, namely those surrounding phrasal verbs and idioms which were analysed in the previous chapter. Lastly, there is the matter of evaluating whether using the BNC was a good choice. Overall, my results points towards it remaining a viable source of data for academic work. However, I must draw attention to the fact that it does contain some anachronistic language that potentially makes it unsuitable for training algorithms that would interact with speakers of modern English. One such example that I came across was the suggestion of ‘compère’ as a paraphrase for “begin the evening”. This is perfectly correct – compère does mean after all to act as a host, or someone who introduces performers in a variety show (OED Online) – and as such it is reasonable for someone to ‘compère the evening’. However, consulting Google’s ngram viewer reveals that ‘compère’ peaked in popularity in 1862 and is no longer as popular as it once was. If nothing else, this underscores the fact that no model is perfect and that any algorithm will only ever be as good as the data it is built upon.\hfill\break\vspace{5mm}

An alternative way of measuring the performance of a system is the phi coefficient. The phi
coefficient describes the quality of a binary classifier by quantifying the correlation between
the predicted labels and the observed data. In fact, it may be a superior measure of performance
than a precision-recall plot since the phi coefficient is better at normalising results when there
are considerable discrepancies between the size of categories. This is the case for this study,
where the number of true negatives (94) is almost twice that of true positives (52). The
following formula defines the phi coefficient (where $P_{T}$ is true positive, $N_{T}$ true negative, $P_{F}$
false positive and $N_{F}$ false negative):\hfill

\begin{gather*}
\phi = \frac{P_{T} \cdot N_{T} - P_{F} \cdot N_{F}}{\sqrt[]{(P_{T} + P_{F})(P_{T} + N_{F})(N_{T} + P_{F})(N_{T} + N_{F})}}\hfill\break
\end{gather*}\vspace{2mm}

This formula returns a value between -1 and +1, where -1 means the predicted and observed data are completely at odds with each other, 0 indicates no better than chance performance and +1 is total agreement between prediction and observation. The phi coefficient for my model trained on word vectors generated using Skip-grams equals \Plus0.61. This indicates a positive correlation between the labels assigned by the classifier and the real values of the data – coherent with human judgement of the paraphrases as either valid or inappropriate. As such, my hypothesis that paraphrasing verbal metonymy using distributional semantics is possible has been confirmed. Furthermore, my experiment agrees with Mikolov et al.’s calculations concerning the performance of the two algorithms introduced by word2vec (2013b).\hfill\break\vspace{5mm}

\chapter{Conclusion}

I have proposed a method with which to model computationally instances of verbal metonymy where an event-selecting verb is combined with an entity-denoting noun. The present study has confirmed that this is an area of language that is ripe for further research using Natural Language Processing. My research has confirmed that word embeddings are a viable way of generating paraphrases for covert events encoded by metonymy. The phi coefficient ($\phi$ = 0.61) for my classifier function indicates a positive correlation between the labels assigned to the data and human judgement of these ranked paraphrases. Additionally, this experiment served to uncover and analyse some of the challenges that lay ahead on the road towards more accurate NLP systems. One such obstacle is generating meaning representations of idioms and phrasal verbs. Ways of refining the current approach with the aim of obtaining a greater degree of accuracy include using more robust dependency parsers and using larger corpora that capture the way people use everyday language. Another way of overcoming this problem is by identifying paraphrases of idioms using the system introduced by Pershina et al. (2015), which could be reimplemented in order to calculate similarity scores as a complement to the model described in the present paper. The second result of my experiment is the confirmation that models trained using the Skip-gram algorithm perform better than those which implement the Continuous bag of words approach. This echoes the findings of Mikolov et al., the authors of both algorithms (2013a; 2013b). Word vectors have proven to be a highly innovative and disruptive technology, yet it is evident that they remain only as good as the corpus they are built from. However, further research that implements word embeddings should not be discouraged, since they provide remarkable results and Mikolov et al. reassure us that “it should be possible to train the CBOW and Skip-gram models even on corpora with one trillion words, for basically unlimited size of the vocabulary” (2013a: 10). An ambitious model which makes use of a corpus of these dimensions would most assuredly see gains in accuracy.\hfill\break\vspace{5mm}

The field of computational linguistics is moving at breakneck speed. Currently ‘Long ShortTerm Memory’ systems – a type of recurrent neural network – threaten word embeddings’ position as the state of the art solution for distributional semantics. One of the best implementations of the LSTM architecture in recent years is Le \& Zuidema’s paper on training a recurrent neural network to perform sentiment analysis (2015). Their solution allows the model to capture long range dependencies and outperforms traditional neural networks (Le \& Zuidema 2015). Their tree-structured LSTM network could be a natural progression away from the Skip-gram model for future research on paraphrasing verbal metonymy. However, there are reasons to be wary of using deep learning to solve certain problems, for the ‘black box’ approach may not always be the best one and there are benefits to having human-interpretable language models. For approaches that continue to use word embeddings, one aspect that can be improved upon is the task of human evaluations of the model. One possible way of doing this is crowdsourcing the evaluation using systems such as Amazon’s Mechanical Turk. Alternatively, performance evaluation could be carried out by using standardised tests. This would make for a simple validation for which human performance is already known. There is precedent for this approach in Rapp (2003) and Turney (2006), who test vector-based representations of meaning on TOEFL and SAT exams and compare their models to the average human score.\hfill\break

\addcontentsline{toc}{chapter}{Appendices}
\chapter*{Appendices}  
\renewcommand{\thesection}{\Roman{section}}

\section{Full set of ranking tables}

This appendix provides the full set of results for the experiment described in this paper. It gives a full account of the paraphrase rankings for the 49 target sentences considered by the model. The ‘Not in vocab.’ label indicates a phrasal verb or idiom (e.g. “Keep the process going” or “Keep an eye on the scene”) that was scraped by algorithm but was unsuccessfully evaluated. Confidence scores with green backgrounds are those above the 0.5 threshold, with grey representing those below 0.5. Additionally, scores preceded by an asterisk and set in bold type are those which human judgement has deemed to be false positives or false negatives. The dataset can be summarised as follows:\\
‘Begin’: 10 instances of verbal metonymy evaluated. 48 paraphrases generated: 12 true
positives and 29 true negatives; 3 false positives and 4 false negatives.\\
‘Enjoy’: 20 instances of verbal metonymy evaluated. 84 paraphrases generated in total: 31 true
positives and 39 true negatives; 9 false positives and 5 false negatives.\\
‘Finish’: 11 instances of verbal metonymy evaluated. 47 paraphrases generated in total: 9 true
positives and 27 true negatives; 2 false positives and 9 false negatives.\hfill\break\vspace{2mm}

Data for verbal metonymy containing ‘begin’:

\renewcommand{\arraystretch}{0.8}
\begin{table}[H]
\centering
\resizebox{\textwidth}{!}{
\begin{tabular}{ccccc}
\multicolumn{2}{l}{begin-1}                                                                                                         & \multicolumn{1}{l}{}  & \multicolumn{2}{l}{begin-2}                                                                                                          \\ \cline{1-2} \cline{4-5} 
\multicolumn{2}{|c|}{\begin{tabular}[c]{@{}c@{}}“Before I began the formal research…”\\ (aca/CRS, sentence 1012)\end{tabular}}      & \multicolumn{1}{c|}{} & \multicolumn{2}{c|}{\begin{tabular}[c]{@{}c@{}}“...he liked to begin the unwinding process.”\\ (fic/CDB, sentence 201)\end{tabular}} \\ \cline{1-2} \cline{4-5} 
\multicolumn{1}{|c}{\cellcolor[HTML]{BBDAFF}Paraphrase candidate} & \multicolumn{1}{c|}{\cellcolor[HTML]{BBDAFF}Confidence}        & \multicolumn{1}{c|}{} & \multicolumn{1}{c}{\cellcolor[HTML]{BBDAFF}Paraphrase candidate}  & \multicolumn{1}{c|}{\cellcolor[HTML]{BBDAFF}Confidence}         \\ 
\multicolumn{1}{|c}{Undertake the research.}                      & \multicolumn{1}{c|}{\cellcolor[HTML]{9AFF99}0.60157}           & \multicolumn{1}{c|}{} & \multicolumn{1}{c}{Undergo the process.}                          & \multicolumn{1}{c|}{\cellcolor[HTML]{9AFF99}0.57620}            \\
\multicolumn{1}{|c}{Conduct the research.}                        & \multicolumn{1}{c|}{\cellcolor[HTML]{9AFF99}0.55156}           & \multicolumn{1}{c|}{} & \multicolumn{1}{c}{Do the process.}                               & \multicolumn{1}{c|}{\cellcolor[HTML]{9AFF99}0.53042}            \\
\multicolumn{1}{|c}{Assist the research.}                         & \multicolumn{1}{c|}{\cellcolor[HTML]{9AFF99}\textbf{*0.53126}} & \multicolumn{1}{c|}{} & \multicolumn{1}{c}{Carry out the process.}                        & \multicolumn{1}{c|}{\cellcolor[HTML]{EFEFEF}\textbf{*0.49178}}  \\ 
\multicolumn{1}{|c}{Inform the research.}                         & \multicolumn{1}{c|}{\cellcolor[HTML]{EFEFEF}0.42233} & \multicolumn{1}{c|}{} & \multicolumn{1}{c}{Build the process.}                        & \multicolumn{1}{c|}{\cellcolor[HTML]{EFEFEF}0.49178}  \\ 
\multicolumn{1}{|c}{Cite the research.}                         & \multicolumn{1}{c|}{\cellcolor[HTML]{EFEFEF}0.32108} & \multicolumn{1}{c|}{} & \multicolumn{1}{c}{Keep the process going.}                        & \multicolumn{1}{c|}{\cellcolor[HTML]{EFEFEF}Not in vocab.}  \\ \cline{4-5} 
\multicolumn{1}{|c}{Set out the research.}                         & \multicolumn{1}{c|}{\cellcolor[HTML]{EFEFEF}Not in vocab.} & \multicolumn{1}{c}{} & \multicolumn{1}{c}{}                        & \multicolumn{1}{c}{}  \\ \cline{1-2}
\end{tabular}
}
\end{table}
\pagestyle{style2}

\renewcommand{\arraystretch}{0.8}
\begin{table}[H]
\centering
\resizebox{\textwidth}{!}{
\begin{tabular}{ccccc}
\multicolumn{2}{l}{begin-3}& \multicolumn{1}{l}{}  & \multicolumn{2}{l}{begin-4}\\ \cline{1-2} \cline{4-5} 
\multicolumn{2}{|c|}{\begin{tabular}[c]{@{}c@{}}
“...any attempt to begin the painful separation…”\\(aca/CTY, sentence 412)
\end{tabular}}      & \multicolumn{1}{c|}{} & \multicolumn{2}{c|}{\begin{tabular}[c]{@{}c@{}}
“I think you should begin the next chapter now.”\\(aca/F9V, sentence 899)
\end{tabular}} \\ \cline{1-2} \cline{4-5} 
\multicolumn{1}{|c}{\cellcolor[HTML]{BBDAFF}Paraphrase candidate} & \multicolumn{1}{c|}{\cellcolor[HTML]{BBDAFF}Confidence}        & \multicolumn{1}{c|}{} & \multicolumn{1}{c}{\cellcolor[HTML]{BBDAFF}Paraphrase candidate}  & \multicolumn{1}{c|}{\cellcolor[HTML]{BBDAFF}Confidence}         \\ 
\multicolumn{1}{|c}{
Reflect the separation.}& \multicolumn{1}{c|}{\cellcolor[HTML]{EFEFEF}0.39124}& \multicolumn{1}{c|}{} & \multicolumn{1}{c}{
Read the chapter.}& \multicolumn{1}{c|}{\cellcolor[HTML]{9AFF99}0.51943}\\\multicolumn{1}{|c}{
Abolish the separation.}& \multicolumn{1}{c|}{\cellcolor[HTML]{EFEFEF}0.38755}& \multicolumn{1}{c|}{} & \multicolumn{1}{c}{
Write the chapter.}& \multicolumn{1}{c|}{\cellcolor[HTML]{EFEFEF}0.35864}\\
\multicolumn{1}{|c}{
Overcome the separation.}& \multicolumn{1}{c|}{\cellcolor[HTML]{EFEFEF}0.38322} & \multicolumn{1}{c|}{} & \multicolumn{1}{c}{Explain the chapter.}& \multicolumn{1}{c|}{\cellcolor[HTML]{EFEFEF}0.35352}  \\ \cline{1-2}
\multicolumn{1}{c}{}& \multicolumn{1}{c}{} & \multicolumn{1}{c|}{} & \multicolumn{1}{c}{Discuss the chapter.}& \multicolumn{1}{c|}{\cellcolor[HTML]{EFEFEF}0.27910}  \\ \cline{4-5} 
\end{tabular}
}
\end{table}

\renewcommand{\arraystretch}{0.8}
\begin{table}[H]
\centering
\resizebox{\textwidth}{!}{
\begin{tabular}{ccccc}
\multicolumn{2}{l}{begin-5}& \multicolumn{1}{l}{}  & \multicolumn{2}{l}{begin-6}\\ \cline{1-2} \cline{4-5} 
\multicolumn{2}{|c|}{\begin{tabular}[c]{@{}c@{}}
“...could've used material from the question\\to begin the essay...”\\(aca/HXH, sentence 1356)
\end{tabular}}      & \multicolumn{1}{c|}{} & \multicolumn{2}{c|}{\begin{tabular}[c]{@{}c@{}}
“...persuaded Louis to begin the task not\\completed…”\\(aca/EA7, sentence 439)
\end{tabular}} \\ \cline{1-2} \cline{4-5} 
\multicolumn{1}{|c}{\cellcolor[HTML]{BBDAFF}Paraphrase candidate} & \multicolumn{1}{c|}{\cellcolor[HTML]{BBDAFF}Confidence}        & \multicolumn{1}{c|}{} & \multicolumn{1}{c}{\cellcolor[HTML]{BBDAFF}Paraphrase candidate}  & \multicolumn{1}{c|}{\cellcolor[HTML]{BBDAFF}Confidence}         \\ 
\multicolumn{1}{|c}{
Start the essay.}& \multicolumn{1}{c|}{\cellcolor[HTML]{9AFF99}\textbf{*0.94397}}& \multicolumn{1}{c|}{} & \multicolumn{1}{c}{
Face the task.}& \multicolumn{1}{c|}{\cellcolor[HTML]{9AFF99}0.59582}\\\multicolumn{1}{|c}{
Write the essay.}& \multicolumn{1}{c|}{\cellcolor[HTML]{9AFF99}0.56490}& \multicolumn{1}{c|}{} & \multicolumn{1}{c}{
Give the task.}& \multicolumn{1}{c|}{\cellcolor[HTML]{EFEFEF}0.43031}\\
\multicolumn{1}{|c}{
Build the essay.}& \multicolumn{1}{c|}{\cellcolor[HTML]{EFEFEF}0.47654} & \multicolumn{1}{c|}{} & \multicolumn{1}{c}{
Allocate the task.}& \multicolumn{1}{c|}{\cellcolor[HTML]{EFEFEF}0.42848}\\
\multicolumn{1}{|c}{
Organise the essay.}& \multicolumn{1}{c|}{\cellcolor[HTML]{EFEFEF}0.44926} & \multicolumn{1}{c|}{} & \multicolumn{1}{c}{
Achieve the task.}& \multicolumn{1}{c|}{\cellcolor[HTML]{EFEFEF}0.39055}\\
\multicolumn{1}{|c}{
Develop the essay.}& \multicolumn{1}{c|}{\cellcolor[HTML]{EFEFEF}\textbf{*0.45284}} & \multicolumn{1}{c|}{} & \multicolumn{1}{c}{
Ignore the task.}& \multicolumn{1}{c|}{\cellcolor[HTML]{EFEFEF}0.34324}\\
\multicolumn{1}{|c}{
Form the essay.}& \multicolumn{1}{c|}{\cellcolor[HTML]{EFEFEF}0.36391} & \multicolumn{1}{c|}{} & \multicolumn{1}{c}{
Delegate the task.}& \multicolumn{1}{c|}{\cellcolor[HTML]{EFEFEF}0.31528}\\
\multicolumn{1}{|c}{
Follow the essay.}& \multicolumn{1}{c|}{\cellcolor[HTML]{EFEFEF}0.35311} & \multicolumn{1}{c|}{} & \multicolumn{1}{c}{
Have the task.}& \multicolumn{1}{c|}{\cellcolor[HTML]{EFEFEF}0.27727}\\
\multicolumn{1}{|c}{
Structure the essay.}& \multicolumn{1}{c|}{\cellcolor[HTML]{EFEFEF}0.34428} & \multicolumn{1}{c|}{} & \multicolumn{1}{c}{
Synthesise the task.}& \multicolumn{1}{c|}{\cellcolor[HTML]{EFEFEF}0.27803}\\
\multicolumn{1}{|c}{
Shape the essay.}& \multicolumn{1}{c|}{\cellcolor[HTML]{EFEFEF}0.33418} & \multicolumn{1}{c|}{} & \multicolumn{1}{c}{
Turn his hand to the task.}& \multicolumn{1}{c|}{\cellcolor[HTML]{EFEFEF}\textbf{*Not in vocab.}}\\\cline{1-2} \cline{4-5} 
\end{tabular}
}
\end{table}

\renewcommand{\arraystretch}{0.8}
\begin{table}[H]
\centering
\resizebox{\textwidth}{!}{
\begin{tabular}{ccccc}
\multicolumn{2}{l}{begin-7}& \multicolumn{1}{l}{}  & \multicolumn{2}{l}{begin-8}\\ \cline{1-2} \cline{4-5} 
\multicolumn{2}{|c|}{\begin{tabular}[c]{@{}c@{}}
“...to begin the usual psalms...”\\(fic/H9C, sentence 2201)
\end{tabular}}      & \multicolumn{1}{c|}{} & \multicolumn{2}{c|}{\begin{tabular}[c]{@{}c@{}}
“...went out to the kitchen to begin\\the dinner.”\\(fic/H9C, sentence 3078)
\end{tabular}} \\ \cline{1-2} \cline{4-5} 
\multicolumn{1}{|c}{\cellcolor[HTML]{BBDAFF}Paraphrase candidate} & \multicolumn{1}{c|}{\cellcolor[HTML]{BBDAFF}Confidence}        & \multicolumn{1}{c|}{} & \multicolumn{1}{c}{\cellcolor[HTML]{BBDAFF}Paraphrase candidate}  & \multicolumn{1}{c|}{\cellcolor[HTML]{BBDAFF}Confidence}         \\ 
\multicolumn{1}{|c}{
Sing the psalm.}& \multicolumn{1}{c|}{\cellcolor[HTML]{9AFF99}0.54752}& \multicolumn{1}{c|}{} & \multicolumn{1}{c}{
Cook the dinner.}& \multicolumn{1}{c|}{\cellcolor[HTML]{9AFF99}0.51848}\\\multicolumn{1}{|c}{
Chant the psalm.}& \multicolumn{1}{c|}{\cellcolor[HTML]{EFEFEF}\textbf{*0.47784}}& \multicolumn{1}{c|}{} & \multicolumn{1}{c}{
Leave the dinner.}& \multicolumn{1}{c|}{\cellcolor[HTML]{EFEFEF}0.33579}\\\cline{1-2} 
\multicolumn{1}{c}{}& \multicolumn{1}{c}{} & \multicolumn{1}{c}{} & \multicolumn{1}{|c}{
Share the dinner.}& \multicolumn{1}{c|}{\cellcolor[HTML]{EFEFEF}0.34324}  \\ \cline{4-5} 
\end{tabular}
}
\end{table}

\renewcommand{\arraystretch}{0.8}
\begin{table}[H]
\centering
\resizebox{\textwidth}{!}{
\begin{tabular}{ccccc}
\multicolumn{2}{l}{begin-9}& \multicolumn{1}{l}{}  & \multicolumn{2}{l}{begin-10}\\ \cline{1-2} \cline{4-5} 
\multicolumn{2}{|c|}{\begin{tabular}[c]{@{}c@{}}
“...Gaveston began the dark satanic ritual...”\\(fic/H9C, sentence 3078)
\end{tabular}}      & \multicolumn{1}{c|}{} & \multicolumn{2}{c|}{\begin{tabular}[c]{@{}c@{}}
“The All Blacks begin the Irish leg of\\their tour...”\\(news/A80, sentence 276)
\end{tabular}} \\ \cline{1-2} \cline{4-5} 
\multicolumn{1}{|c}{\cellcolor[HTML]{BBDAFF}Paraphrase candidate} & \multicolumn{1}{c|}{\cellcolor[HTML]{BBDAFF}Confidence}        & \multicolumn{1}{c|}{} & \multicolumn{1}{c}{\cellcolor[HTML]{BBDAFF}Paraphrase candidate}  & \multicolumn{1}{c|}{\cellcolor[HTML]{BBDAFF}Confidence}         \\ 
\multicolumn{1}{|c}{
Perform the ritual.}& \multicolumn{1}{c|}{\cellcolor[HTML]{9AFF99}0.67040}& \multicolumn{1}{c|}{} & \multicolumn{1}{c}{
Go on the leg.}& \multicolumn{1}{c|}{\cellcolor[HTML]{9AFF99}0.60371}\\\multicolumn{1}{|c}{
Complete the ritual.}& \multicolumn{1}{c|}{\cellcolor[HTML]{9AFF99}\textbf{*0.54511}}& \multicolumn{1}{c|}{} & \multicolumn{1}{c}{
Prepare for the leg.}& \multicolumn{1}{c|}{\cellcolor[HTML]{9AFF99}0.59806}\\\multicolumn{1}{|c}{
Witness the ritual.}& \multicolumn{1}{c|}{\cellcolor[HTML]{EFEFEF}0.35892}& \multicolumn{1}{c|}{} & \multicolumn{1}{c}{
Face the leg.}& \multicolumn{1}{c|}{\cellcolor[HTML]{EFEFEF}\textbf{*0.49084}}\\\cline{1-2}\multicolumn{1}{c}{}& \multicolumn{1}{c}{}& \multicolumn{1}{c}{} & \multicolumn{1}{|c}{
Win the leg.}& \multicolumn{1}{c|}{\cellcolor[HTML]{EFEFEF}0.33879}\\\cline{4-5} 
\end{tabular}
}
\end{table}

\vspace{2mm}
Data for verbal metonymy containing ‘enjoy’:\hfill\break

\renewcommand{\arraystretch}{0.8}
\begin{table}[H]
\centering
\resizebox{\textwidth}{!}{
\begin{tabular}{ccccc}
\multicolumn{2}{l}{enjoy-1}& \multicolumn{1}{l}{}  & \multicolumn{2}{l}{enjoy-2}\\ \cline{1-2} \cline{4-5} 
\multicolumn{2}{|c|}{\begin{tabular}[c]{@{}c@{}}
“...the union enjoys the same defences as\\an individual.”\\(aca/FSS, sentence 1456)
\end{tabular}}      & \multicolumn{1}{c|}{} & \multicolumn{2}{c|}{\begin{tabular}[c]{@{}c@{}}
“...a coalition would enjoy the support...”\\(aca/J57, sentence 1719)
\end{tabular}} \\ \cline{1-2} \cline{4-5} 
\multicolumn{1}{|c}{\cellcolor[HTML]{BBDAFF}Paraphrase candidate} & \multicolumn{1}{c|}{\cellcolor[HTML]{BBDAFF}Confidence}        & \multicolumn{1}{c|}{} & \multicolumn{1}{c}{\cellcolor[HTML]{BBDAFF}Paraphrase candidate}  & \multicolumn{1}{c|}{\cellcolor[HTML]{BBDAFF}Confidence}         \\ 
\multicolumn{1}{|c}{
Have the defence.}& \multicolumn{1}{c|}{\cellcolor[HTML]{9AFF99}0.53353}& \multicolumn{1}{c|}{} & \multicolumn{1}{c}{
Receive the support.}& \multicolumn{1}{c|}{\cellcolor[HTML]{9AFF99}0.59601}\\\multicolumn{1}{|c}{
Apply the defence.}& \multicolumn{1}{c|}{\cellcolor[HTML]{EFEFEF}0.49698}& \multicolumn{1}{c|}{} & \multicolumn{1}{c}{
Have the support.}& \multicolumn{1}{c|}{\cellcolor[HTML]{9AFF99}0.52088}\\\multicolumn{1}{|c}{
Pursue the defence.}& \multicolumn{1}{c|}{\cellcolor[HTML]{EFEFEF}0.47940}& \multicolumn{1}{c|}{} & \multicolumn{1}{c}{
Secure the support.}& \multicolumn{1}{c|}{\cellcolor[HTML]{EFEFEF}\textbf{*0.48592}}\\\multicolumn{1}{|c}{
Assess the defence.}& \multicolumn{1}{c|}{\cellcolor[HTML]{EFEFEF}0.37605}& \multicolumn{1}{c|}{} & \multicolumn{1}{c}{
Rally the support.}& \multicolumn{1}{c|}{\cellcolor[HTML]{EFEFEF}0.33238}\\\multicolumn{1}{|c}{
Contest the defence.}& \multicolumn{1}{c|}{\cellcolor[HTML]{EFEFEF}0.35429}& \multicolumn{1}{c|}{} & \multicolumn{1}{c}{
Command the support.}& \multicolumn{1}{c|}{\cellcolor[HTML]{EFEFEF}0.28217}\\\cline{1-2} \cline{4-5} 
\end{tabular}
}
\end{table}

\renewcommand{\arraystretch}{0.8}
\begin{table}[H]
\centering
\resizebox{\textwidth}{!}{
\begin{tabular}{ccccc}
\multicolumn{2}{l}{enjoy-3}& \multicolumn{1}{l}{}  & \multicolumn{2}{l}{enjoy-4}\\ \cline{1-2} \cline{4-5} 
\multicolumn{2}{|c|}{\begin{tabular}[c]{@{}c@{}}
“He seems to enjoy the job doesn't he?”\\(dem/KPB, sentence 2187)
\end{tabular}}      & \multicolumn{1}{c|}{} & \multicolumn{2}{c|}{\begin{tabular}[c]{@{}c@{}}
“...well erm I enjoyed the Mozart…”\\(dem/KPU, sentence 955)
\end{tabular}} \\ \cline{1-2} \cline{4-5} 
\multicolumn{1}{|c}{\cellcolor[HTML]{BBDAFF}Paraphrase candidate} & \multicolumn{1}{c|}{\cellcolor[HTML]{BBDAFF}Confidence}        & \multicolumn{1}{c|}{} & \multicolumn{1}{c}{\cellcolor[HTML]{BBDAFF}Paraphrase candidate}  & \multicolumn{1}{c|}{\cellcolor[HTML]{BBDAFF}Confidence}         \\ 
\multicolumn{1}{|c}{
Do the job.}& \multicolumn{1}{c|}{\cellcolor[HTML]{9AFF99}0.68356}& \multicolumn{1}{c|}{} & \multicolumn{1}{c}{
Listen to the Mozart.}& \multicolumn{1}{c|}{\cellcolor[HTML]{9AFF99}0.67809}\\\multicolumn{1}{|c}{
Get the job.}& \multicolumn{1}{c|}{\cellcolor[HTML]{9AFF99}\textbf{*0.65040}}& \multicolumn{1}{c|}{} & \multicolumn{1}{c}{
See the Mozart.}& \multicolumn{1}{c|}{\cellcolor[HTML]{9AFF99}\textbf{*0.59159}}\\\multicolumn{1}{|c}{
See the job.}& \multicolumn{1}{c|}{\cellcolor[HTML]{EFEFEF}0.32827}& \multicolumn{1}{c|}{} & \multicolumn{1}{c}{
Hear the Mozart.}& \multicolumn{1}{c|}{\cellcolor[HTML]{9AFF99}0.53761}\\\cline{1-2} \cline{4-5} 
\end{tabular}
}
\end{table}

\renewcommand{\arraystretch}{0.8}
\begin{table}[H]
\centering
\resizebox{\textwidth}{!}{
\begin{tabular}{ccccc}
\multicolumn{2}{l}{enjoy-5}& \multicolumn{1}{l}{}  & \multicolumn{2}{l}{enjoy-6}\\ \cline{1-2} \cline{4-5} 
\multicolumn{2}{|c|}{\begin{tabular}[c]{@{}c@{}}
“You'll enjoy the story.”\\(dem/KBW, sentence 9489)
\end{tabular}}      & \multicolumn{1}{c|}{} & \multicolumn{2}{c|}{\begin{tabular}[c]{@{}c@{}}
“Though I enjoyed the book immensely…”\\(news/K37, sentence 174)
\end{tabular}} \\ \cline{1-2} \cline{4-5} 
\multicolumn{1}{|c}{\cellcolor[HTML]{BBDAFF}Paraphrase candidate} & \multicolumn{1}{c|}{\cellcolor[HTML]{BBDAFF}Confidence}        & \multicolumn{1}{c|}{} & \multicolumn{1}{c}{\cellcolor[HTML]{BBDAFF}Paraphrase candidate}  & \multicolumn{1}{c|}{\cellcolor[HTML]{BBDAFF}Confidence}         \\ 
\multicolumn{1}{|c}{
Hear the story.}& \multicolumn{1}{c|}{\cellcolor[HTML]{9AFF99}\textbf{*0.61463}}& \multicolumn{1}{c|}{} & \multicolumn{1}{c}{
Read the book.}& \multicolumn{1}{c|}{\cellcolor[HTML]{9AFF99}\textbf{*0.53622}}\\\multicolumn{1}{|c}{
Read the story.}& \multicolumn{1}{c|}{\cellcolor[HTML]{9AFF99}0.55733}& \multicolumn{1}{c|}{} & \multicolumn{1}{c}{
Delve (into) the book.}& \multicolumn{1}{c|}{\cellcolor[HTML]{EFEFEF}0.48435}\\\multicolumn{1}{|c}{
Know the story.}& \multicolumn{1}{c|}{\cellcolor[HTML]{9AFF99}0.53276}& \multicolumn{1}{c|}{} & \multicolumn{1}{c}{
Sell the book.}& \multicolumn{1}{c|}{\cellcolor[HTML]{EFEFEF}0.43931}\\\multicolumn{1}{|c}{
Tell the story.}& \multicolumn{1}{c|}{\cellcolor[HTML]{EFEFEF}0.48357}& \multicolumn{1}{c|}{} & \multicolumn{1}{c}{
Publish the book.}& \multicolumn{1}{c|}{\cellcolor[HTML]{EFEFEF}0.42264}\\\multicolumn{1}{|c}{
Like the story.}& \multicolumn{1}{c|}{\cellcolor[HTML]{EFEFEF}0.48325}& \multicolumn{1}{c|}{} & \multicolumn{1}{c}{
Review the book.}& \multicolumn{1}{c|}{\cellcolor[HTML]{EFEFEF}0.34899}\\\multicolumn{1}{|c}{
Finish the story.}& \multicolumn{1}{c|}{\cellcolor[HTML]{EFEFEF}0.43046}& \multicolumn{1}{c|}{} & \multicolumn{1}{c}{
Research the book.}& \multicolumn{1}{c|}{\cellcolor[HTML]{EFEFEF}0.26940}\\\cline{1-2} \cline{4-5} 
\end{tabular}
}
\end{table}

\renewcommand{\arraystretch}{0.8}
\begin{table}[H]
\centering
\resizebox{\textwidth}{!}{
\begin{tabular}{ccccc}
\multicolumn{2}{l}{enjoy-7}& \multicolumn{1}{l}{}  & \multicolumn{2}{l}{enjoy-8}\\ \cline{1-2} \cline{4-5} 
\multicolumn{2}{|c|}{\begin{tabular}[c]{@{}c@{}}
“I've enjoyed the concert but…”\\(dem/KPU, sentence 975)
\end{tabular}}      & \multicolumn{1}{c|}{} & \multicolumn{2}{c|}{\begin{tabular}[c]{@{}c@{}}
“Charlotte did not enjoy the journey…”\\(fic/CB5, sentence 1908)
\end{tabular}} \\ \cline{1-2} \cline{4-5} 
\multicolumn{1}{|c}{\cellcolor[HTML]{BBDAFF}Paraphrase candidate} & \multicolumn{1}{c|}{\cellcolor[HTML]{BBDAFF}Confidence}        & \multicolumn{1}{c|}{} & \multicolumn{1}{c}{\cellcolor[HTML]{BBDAFF}Paraphrase candidate}  & \multicolumn{1}{c|}{\cellcolor[HTML]{BBDAFF}Confidence}         \\ 
\multicolumn{1}{|c}{
See concert.}& \multicolumn{1}{c|}{\cellcolor[HTML]{9AFF99}0.68158}& \multicolumn{1}{c|}{} & \multicolumn{1}{c}{
Spend the journey.}& \multicolumn{1}{c|}{\cellcolor[HTML]{9AFF99}\textbf{*0.64580}}\\\multicolumn{1}{|c}{
Listen to the concert.}& \multicolumn{1}{c|}{\cellcolor[HTML]{9AFF99}0.58792}& \multicolumn{1}{c|}{} & \multicolumn{1}{c}{
Observe the journey.}& \multicolumn{1}{c|}{\cellcolor[HTML]{EFEFEF}0.47043}\\\multicolumn{1}{|c}{
Go to the concert.}& \multicolumn{1}{c|}{\cellcolor[HTML]{9AFF99}0.55673}& \multicolumn{1}{c|}{} & \multicolumn{1}{c}{
Complete the journey.}& \multicolumn{1}{c|}{\cellcolor[HTML]{EFEFEF}0.49725}\\\cline{1-2} \cline{4-5} 
\end{tabular}
}
\end{table}

\renewcommand{\arraystretch}{0.8}
\begin{table}[H]
\centering
\resizebox{\textwidth}{!}{
\begin{tabular}{ccccc}
\multicolumn{2}{l}{enjoy-9}& \multicolumn{1}{l}{}  & \multicolumn{2}{l}{enjoy-10}\\ \cline{1-2} \cline{4-5} 
\multicolumn{2}{|c|}{\begin{tabular}[c]{@{}c@{}}
“Owen himself rather enjoyed the view…”\\(fic/J10, sentence 946)
\end{tabular}}      & \multicolumn{1}{c|}{} & \multicolumn{2}{c|}{\begin{tabular}[c]{@{}c@{}}
“...doing the job by myself and enjoying\\the work.”\\(fic/CCW, sentence 2133)
\end{tabular}} \\ \cline{1-2} \cline{4-5} 
\multicolumn{1}{|c}{\cellcolor[HTML]{BBDAFF}Paraphrase candidate} & \multicolumn{1}{c|}{\cellcolor[HTML]{BBDAFF}Confidence}        & \multicolumn{1}{c|}{} & \multicolumn{1}{c}{\cellcolor[HTML]{BBDAFF}Paraphrase candidate}  & \multicolumn{1}{c|}{\cellcolor[HTML]{BBDAFF}Confidence}         \\ 
\multicolumn{1}{|c}{
See the view.}& \multicolumn{1}{c|}{\cellcolor[HTML]{9AFF99}0.67598}& \multicolumn{1}{c|}{} & \multicolumn{1}{c}{
Do the work.}& \multicolumn{1}{c|}{\cellcolor[HTML]{9AFF99}0.62705}\\\multicolumn{1}{|c}{
Admire the view.}& \multicolumn{1}{c|}{\cellcolor[HTML]{9AFF99}0.67346}& \multicolumn{1}{c|}{} & \multicolumn{1}{c}{
Continue the work.}& \multicolumn{1}{c|}{\cellcolor[HTML]{9AFF99}0.54503}\\\multicolumn{1}{|c}{
Look at the view.}& \multicolumn{1}{c|}{\cellcolor[HTML]{9AFF99}0.52113}& \multicolumn{1}{c|}{} & \multicolumn{1}{c}{
Handle the work.}& \multicolumn{1}{c|}{\cellcolor[HTML]{EFEFEF}0.42257}\\\multicolumn{1}{|c}{
Screen the view.}& \multicolumn{1}{c|}{\cellcolor[HTML]{EFEFEF}0.36740}& \multicolumn{1}{c|}{} & \multicolumn{1}{c}{
Grudge the work.}& \multicolumn{1}{c|}{\cellcolor[HTML]{EFEFEF}0.34002}\\\cline{1-2}
\multicolumn{1}{c}{}& \multicolumn{1}{c}{}& \multicolumn{1}{c|}{} & \multicolumn{1}{c}{
Delay the work.}& \multicolumn{1}{c|}{\cellcolor[HTML]{EFEFEF}0.30198}\\\cline{4-5} 
\end{tabular}
}
\end{table}

\renewcommand{\arraystretch}{0.8}
\begin{table}[H]
\centering
\resizebox{\textwidth}{!}{
\begin{tabular}{ccccc}
\multicolumn{2}{l}{enjoy-11}& \multicolumn{1}{l}{}  & \multicolumn{2}{l}{enjoy-12}\\ \cline{1-2} \cline{4-5} 
\multicolumn{2}{|c|}{\begin{tabular}[c]{@{}c@{}}
“She enjoyed the opera…”\\(fic/G0Y, sentence 2445)
\end{tabular}}      & \multicolumn{1}{c|}{} & \multicolumn{2}{c|}{\begin{tabular}[c]{@{}c@{}}
“...anyone who enjoys the criticism.”\\(news/AHC, sentence 342)
\end{tabular}} \\ \cline{1-2} \cline{4-5} 
\multicolumn{1}{|c}{\cellcolor[HTML]{BBDAFF}Paraphrase candidate} & \multicolumn{1}{c|}{\cellcolor[HTML]{BBDAFF}Confidence}        & \multicolumn{1}{c|}{} & \multicolumn{1}{c}{\cellcolor[HTML]{BBDAFF}Paraphrase candidate}  & \multicolumn{1}{c|}{\cellcolor[HTML]{BBDAFF}Confidence}         \\ 
\multicolumn{1}{|c}{
Go to the opera.}& \multicolumn{1}{c|}{\cellcolor[HTML]{9AFF99}0.56013}& \multicolumn{1}{c|}{} & \multicolumn{1}{c}{
Take the criticism.}& \multicolumn{1}{c|}{\cellcolor[HTML]{9AFF99}0.60199}\\\multicolumn{1}{|c}{
Visit the opera.}& \multicolumn{1}{c|}{\cellcolor[HTML]{9AFF99}0.51182}& \multicolumn{1}{c|}{} & \multicolumn{1}{c}{
Come under criticism.}& \multicolumn{1}{c|}{\cellcolor[HTML]{9AFF99}0.58639}\\ \cline{1-2} \multicolumn{1}{c}{}& \multicolumn{1}{c}{}& \multicolumn{1}{c|}{} & \multicolumn{1}{c}{
Accept the criticism.}& \multicolumn{1}{c|}{\cellcolor[HTML]{9AFF99}0.51384}\\\multicolumn{1}{c}{}& \multicolumn{1}{c}{}& \multicolumn{1}{c|}{} & \multicolumn{1}{c}{
Face the criticism.}& \multicolumn{1}{c|}{\cellcolor[HTML]{EFEFEF}\textbf{*0.38531}}\\\multicolumn{1}{c}{}& \multicolumn{1}{c}{}& \multicolumn{1}{c|}{} & \multicolumn{1}{c}{
Rebut the criticism.}& \multicolumn{1}{c|}{\cellcolor[HTML]{EFEFEF}0.36010}\\\cline{4-5} 
\end{tabular}
}
\end{table}

\renewcommand{\arraystretch}{0.8}
\begin{table}[H]
\centering
\resizebox{\textwidth}{!}{
\begin{tabular}{ccccc}
\multicolumn{2}{l}{enjoy-13}& \multicolumn{1}{l}{}  & \multicolumn{2}{l}{enjoy-14}\\ \cline{1-2} \cline{4-5} 
\multicolumn{2}{|c|}{\begin{tabular}[c]{@{}c@{}}
“You will not enjoy the meeting.”\\(fic/H85, sentence 1150)
\end{tabular}}      & \multicolumn{1}{c|}{} & \multicolumn{2}{c|}{\begin{tabular}[c]{@{}c@{}}
“If only Kit could enjoy the scene…”\\(fic/G0S, sentence 1968)
\end{tabular}} \\ \cline{1-2} \cline{4-5} 
\multicolumn{1}{|c}{\cellcolor[HTML]{BBDAFF}Paraphrase candidate} & \multicolumn{1}{c|}{\cellcolor[HTML]{BBDAFF}Confidence}        & \multicolumn{1}{c|}{} & \multicolumn{1}{c}{\cellcolor[HTML]{BBDAFF}Paraphrase candidate}  & \multicolumn{1}{c|}{\cellcolor[HTML]{BBDAFF}Confidence}         \\ 
\multicolumn{1}{|c}{
Attend the meeting.}& \multicolumn{1}{c|}{\cellcolor[HTML]{9AFF99}0.54394}& \multicolumn{1}{c|}{} & \multicolumn{1}{c}{
See the scene.}& \multicolumn{1}{c|}{\cellcolor[HTML]{9AFF99}0.68378}\\\multicolumn{1}{|c}{
Ensure the meeting.}& \multicolumn{1}{c|}{\cellcolor[HTML]{9AFF99}\textbf{*0.50981}}& \multicolumn{1}{c|}{} & \multicolumn{1}{c}{
Watch the scene.}& \multicolumn{1}{c|}{\cellcolor[HTML]{9AFF99}0.67125}\\\multicolumn{1}{|c}{
Arrange the meeting.}& \multicolumn{1}{c|}{\cellcolor[HTML]{EFEFEF}0.48937}& \multicolumn{1}{c|}{} & \multicolumn{1}{c}{
Take in the scene.}& \multicolumn{1}{c|}{\cellcolor[HTML]{9AFF99}0.59520}\\\multicolumn{1}{|c}{
Open the meeting.}& \multicolumn{1}{c|}{\cellcolor[HTML]{EFEFEF}0.43172}& \multicolumn{1}{c|}{} & \multicolumn{1}{c}{
Visualise the scene.}& \multicolumn{1}{c|}{\cellcolor[HTML]{EFEFEF}0.48953}\\\multicolumn{1}{|c}{
Witness the meeting.}& \multicolumn{1}{c|}{\cellcolor[HTML]{EFEFEF}0.39758}& \multicolumn{1}{c|}{} & \multicolumn{1}{c}{
Stare at the scene.}& \multicolumn{1}{c|}{\cellcolor[HTML]{EFEFEF}0.42828}\\\multicolumn{1}{|c}{
End the meeting.}& \multicolumn{1}{c|}{\cellcolor[HTML]{EFEFEF}0.38716}& \multicolumn{1}{c|}{} & \multicolumn{1}{c}{
Picture the scene.}& \multicolumn{1}{c|}{\cellcolor[HTML]{EFEFEF}0.34204}\\\multicolumn{1}{|c}{
Chair the meeting.}& \multicolumn{1}{c|}{\cellcolor[HTML]{EFEFEF}0.33628}& \multicolumn{1}{c|}{} & \multicolumn{1}{c}{
Survey the scene.}& \multicolumn{1}{c|}{\cellcolor[HTML]{EFEFEF}\textbf{*0.29907}}\\\cline{1-2} \multicolumn{1}{c}{}& \multicolumn{1}{c}{}& \multicolumn{1}{c|}{} & \multicolumn{1}{c}{
Keep an eye on the scene.}& \multicolumn{1}{c|}{\cellcolor[HTML]{EFEFEF}Not in vocab.}\\\cline{4-5} 
\end{tabular}
}
\end{table}

\renewcommand{\arraystretch}{0.8}
\begin{table}[H]
\centering
\resizebox{\textwidth}{!}{
\begin{tabular}{ccccc}
\multicolumn{2}{l}{enjoy-15}& \multicolumn{1}{l}{}  & \multicolumn{2}{l}{enjoy-16}\\ \cline{1-2} \cline{4-5} 
\multicolumn{2}{|c|}{\begin{tabular}[c]{@{}c@{}}
“…she enjoyed the smell and the\\sound of them.”\\(fic/J54, sentence 1218)
\end{tabular}}      & \multicolumn{1}{c|}{} & \multicolumn{2}{c|}{\begin{tabular}[c]{@{}c@{}}
“...the professionals enjoying\\the advantages...”\\(news/A8P, sentence 87)
\end{tabular}} \\ \cline{1-2} \cline{4-5} 
\multicolumn{1}{|c}{\cellcolor[HTML]{BBDAFF}Paraphrase candidate} & \multicolumn{1}{c|}{\cellcolor[HTML]{BBDAFF}Confidence}        & \multicolumn{1}{c|}{} & \multicolumn{1}{c}{\cellcolor[HTML]{BBDAFF}Paraphrase candidate}  & \multicolumn{1}{c|}{\cellcolor[HTML]{BBDAFF}Confidence}         \\ 
\multicolumn{1}{|c}{
Inhale the smell.}& \multicolumn{1}{c|}{\cellcolor[HTML]{9AFF99}0.51965}& \multicolumn{1}{c|}{} & \multicolumn{1}{c}{
Have the advantage.}& \multicolumn{1}{c|}{\cellcolor[HTML]{EFEFEF}\textbf{*0.48953}}\\\multicolumn{1}{|c}{
Catch the smell.}& \multicolumn{1}{c|}{\cellcolor[HTML]{9AFF99}0.50337}& \multicolumn{1}{c|}{} & \multicolumn{1}{c}{
Is the advantage.}& \multicolumn{1}{c|}{\cellcolor[HTML]{EFEFEF}0.47691}\\\multicolumn{1}{|c}{
Notice the smell.}& \multicolumn{1}{c|}{\cellcolor[HTML]{EFEFEF}0.38023}& \multicolumn{1}{c|}{} & \multicolumn{1}{c}{
Explain the advantage.}& \multicolumn{1}{c|}{\cellcolor[HTML]{EFEFEF}0.38402}\\\cline{4-5} \multicolumn{1}{|c}{
Detect the smell.}& \multicolumn{1}{c|}{\cellcolor[HTML]{EFEFEF}0.31561}& \multicolumn{1}{c}{} & \multicolumn{1}{c}{}& \multicolumn{1}{c}{}\\\cline{1-2}
\end{tabular}
}
\end{table}

\renewcommand{\arraystretch}{0.8}
\begin{table}[H]
\centering
\resizebox{\textwidth}{!}{
\begin{tabular}{ccccc}
\multicolumn{2}{l}{enjoy-17}& \multicolumn{1}{l}{}  & \multicolumn{2}{l}{enjoy-18}\\ \cline{1-2} \cline{4-5} 
\multicolumn{2}{|c|}{\begin{tabular}[c]{@{}c@{}}
“...Austria traditionally enjoys\\the distinction…”\\(news/A3P, sentence 109)
\end{tabular}}      & \multicolumn{1}{c|}{} & \multicolumn{2}{c|}{\begin{tabular}[c]{@{}c@{}}
“...appears to enjoy the attentions of\\his doting...”\\(news/K37, sentence 188)
\end{tabular}} \\ \cline{1-2} \cline{4-5} 
\multicolumn{1}{|c}{\cellcolor[HTML]{BBDAFF}Paraphrase candidate} & \multicolumn{1}{c|}{\cellcolor[HTML]{BBDAFF}Confidence}        & \multicolumn{1}{c|}{} & \multicolumn{1}{c}{\cellcolor[HTML]{BBDAFF}Paraphrase candidate}  & \multicolumn{1}{c|}{\cellcolor[HTML]{BBDAFF}Confidence}         \\ 
\multicolumn{1}{|c}{
Have the distinction.}& \multicolumn{1}{c|}{\cellcolor[HTML]{9AFF99}0.61792}& \multicolumn{1}{c|}{} & \multicolumn{1}{c}{
Have the attention. }& \multicolumn{1}{c|}{\cellcolor[HTML]{9AFF99}0.62154}\\\multicolumn{1}{|c}{
Make the distinction.}& \multicolumn{1}{c|}{\cellcolor[HTML]{9AFF99}\textbf{*0.52451}}& \multicolumn{1}{c|}{} & \multicolumn{1}{c}{
Attract the attention. }& \multicolumn{1}{c|}{\cellcolor[HTML]{9AFF99}0.55481}\\\multicolumn{1}{|c}{
Give the distinction.}& \multicolumn{1}{c|}{\cellcolor[HTML]{EFEFEF}0.49339}& \multicolumn{1}{c|}{} & \multicolumn{1}{c}{
Catch the attention.}& \multicolumn{1}{c|}{\cellcolor[HTML]{9AFF99}0.50076}\\\cline{1-2} \cline{4-5} 
\end{tabular}
}
\end{table}

\renewcommand{\arraystretch}{0.8}
\begin{table}[H]
\centering
\resizebox{\textwidth}{!}{
\begin{tabular}{ccccc}
\multicolumn{2}{l}{enjoy-19}& \multicolumn{1}{l}{}  & \multicolumn{2}{l}{enjoy-20}\\ \cline{1-2} \cline{4-5} 
\multicolumn{2}{|c|}{\begin{tabular}[c]{@{}c@{}}
“...we set forth to enjoy the countryside.”\\(news/AJF, sentence 255)
\end{tabular}}      & \multicolumn{1}{c|}{} & \multicolumn{2}{c|}{\begin{tabular}[c]{@{}c@{}}
“…say I actually enjoyed the experience…”\\(news/AHC, sentence 741)
\end{tabular}} \\ \cline{1-2} \cline{4-5} 
\multicolumn{1}{|c}{\cellcolor[HTML]{BBDAFF}Paraphrase candidate} & \multicolumn{1}{c|}{\cellcolor[HTML]{BBDAFF}Confidence}        & \multicolumn{1}{c|}{} & \multicolumn{1}{c}{\cellcolor[HTML]{BBDAFF}Paraphrase candidate}  & \multicolumn{1}{c|}{\cellcolor[HTML]{BBDAFF}Confidence}         \\ 
\multicolumn{1}{|c}{
Love the countryside.}& \multicolumn{1}{c|}{\cellcolor[HTML]{9AFF99}\textbf{*0.68629}}& \multicolumn{1}{c|}{} & \multicolumn{1}{c}{
Appreciate the experience.}& \multicolumn{1}{c|}{\cellcolor[HTML]{9AFF99}\textbf{*0.83557}}\\\multicolumn{1}{|c}{
Visit the countryside.}& \multicolumn{1}{c|}{\cellcolor[HTML]{EFEFEF}\textbf{*0.41460}}& \multicolumn{1}{c|}{} & \multicolumn{1}{c}{
Have the experience.}& \multicolumn{1}{c|}{\cellcolor[HTML]{9AFF99}0.67259}\\\multicolumn{1}{|c}{
Scour the countryside.}& \multicolumn{1}{c|}{\cellcolor[HTML]{EFEFEF}0.34624}& \multicolumn{1}{c|}{} & \multicolumn{1}{c}{
Forget the experience.}& \multicolumn{1}{c|}{\cellcolor[HTML]{EFEFEF}0.49317}\\\cline{1-2} \cline{4-5} 
\end{tabular}
}
\end{table}

\vspace{2mm}
Data for verbal metonymy containing ‘finish’:\hfill\break

\renewcommand{\arraystretch}{0.8}
\begin{table}[H]
\centering
\resizebox{\textwidth}{!}{
\begin{tabular}{ccccc}
\multicolumn{2}{l}{finish-1}& \multicolumn{1}{l}{}  & \multicolumn{2}{l}{finish-2}\\ \cline{1-2} \cline{4-5} 
\multicolumn{2}{|c|}{\begin{tabular}[c]{@{}c@{}}
“…we haven't finished the garden.”\\(dem/KP5, sentence 851)
\end{tabular}}      & \multicolumn{1}{c|}{} & \multicolumn{2}{c|}{\begin{tabular}[c]{@{}c@{}}
“You haven't finished the work\\over there…”\\(dem/KBW, sentence 16021)
\end{tabular}} \\ \cline{1-2} \cline{4-5} 
\multicolumn{1}{|c}{\cellcolor[HTML]{BBDAFF}Paraphrase candidate} & \multicolumn{1}{c|}{\cellcolor[HTML]{BBDAFF}Confidence}        & \multicolumn{1}{c|}{} & \multicolumn{1}{c}{\cellcolor[HTML]{BBDAFF}Paraphrase candidate}  & \multicolumn{1}{c|}{\cellcolor[HTML]{BBDAFF}Confidence}         \\ 
\multicolumn{1}{|c}{
Do the garden.}& \multicolumn{1}{c|}{\cellcolor[HTML]{9AFF99}0.57878}& \multicolumn{1}{c|}{} & \multicolumn{1}{c}{
Do the work.}& \multicolumn{1}{c|}{\cellcolor[HTML]{9AFF99}0.57858}\\\multicolumn{1}{|c}{
Go (in) the garden.}& \multicolumn{1}{c|}{\cellcolor[HTML]{EFEFEF}0.48242}& \multicolumn{1}{c|}{} & \multicolumn{1}{c}{
Get the work.}& \multicolumn{1}{c|}{\cellcolor[HTML]{EFEFEF}0.47965}\\\multicolumn{1}{|c}{
Dig the garden.}& \multicolumn{1}{c|}{\cellcolor[HTML]{EFEFEF}\textbf{*0.35942}}& \multicolumn{1}{c}{} & \multicolumn{1}{|c}{Carry (on with) the work.}& \multicolumn{1}{c|}{\cellcolor[HTML]{EFEFEF}0.43916}\\\cline{1-2} \cline{4-5}
\end{tabular}
}
\end{table}

\renewcommand{\arraystretch}{0.8}
\begin{table}[H]
\centering
\resizebox{\textwidth}{!}{
\begin{tabular}{ccccc}
\multicolumn{2}{l}{finish-3}& \multicolumn{1}{l}{}  & \multicolumn{2}{l}{finish-4}\\ \cline{1-2} \cline{4-5} 
\multicolumn{2}{|c|}{\begin{tabular}[c]{@{}c@{}}
“I won't finish the whole book.”\\(dem/KBW, sentence 17355)\\
\end{tabular}}      & \multicolumn{1}{c|}{} & \multicolumn{2}{c|}{\begin{tabular}[c]{@{}c@{}}
“And then we finished the game…”\\(dem/KB7, sentence 501)\\
\end{tabular}} \\ \cline{1-2} \cline{4-5} 
\multicolumn{1}{|c}{\cellcolor[HTML]{BBDAFF}Paraphrase candidate} & \multicolumn{1}{c|}{\cellcolor[HTML]{BBDAFF}Confidence}        & \multicolumn{1}{c|}{} & \multicolumn{1}{c}{\cellcolor[HTML]{BBDAFF}Paraphrase candidate}  & \multicolumn{1}{c|}{\cellcolor[HTML]{BBDAFF}Confidence}         \\ 
\multicolumn{1}{|c}{
Read the book.}& \multicolumn{1}{c|}{\cellcolor[HTML]{9AFF99}0.59457}& \multicolumn{1}{c|}{} & \multicolumn{1}{c}{
Win the game.}& \multicolumn{1}{c|}{\cellcolor[HTML]{9AFF99}0.8159}\\\multicolumn{1}{|c}{
Put the book.}& \multicolumn{1}{c|}{\cellcolor[HTML]{EFEFEF}0.47582}& \multicolumn{1}{c|}{} & \multicolumn{1}{c}{
End the game.}& \multicolumn{1}{c|}{\cellcolor[HTML]{9AFF99}\textbf{*0.59684}}\\\multicolumn{1}{|c}{
See the book.}& \multicolumn{1}{c|}{\cellcolor[HTML]{EFEFEF}0.46215}& \multicolumn{1}{c|}{} & \multicolumn{1}{c}{
Play the game.}& \multicolumn{1}{c|}{\cellcolor[HTML]{9AFF99}0.56021}\\\multicolumn{1}{|c}{
Bring the book.}& \multicolumn{1}{c|}{\cellcolor[HTML]{EFEFEF}0.37519}& \multicolumn{1}{c|}{} & \multicolumn{1}{c}{
Buy the game.}& \multicolumn{1}{c|}{\cellcolor[HTML]{EFEFEF}0.39317}\\\multicolumn{1}{|c}{
Have the book.}& \multicolumn{1}{c|}{\cellcolor[HTML]{EFEFEF}0.37220}& \multicolumn{1}{c|}{} & \multicolumn{1}{c}{
Like the game.}& \multicolumn{1}{c|}{\cellcolor[HTML]{EFEFEF}0.33214}\\\cline{4-5} \multicolumn{1}{|c}{
Wade through the book.}& \multicolumn{1}{c|}{\cellcolor[HTML]{EFEFEF}Not in vocab.}& \multicolumn{1}{c}{} & \multicolumn{1}{c}{}& \multicolumn{1}{c}{}\\\cline{1-2}
\end{tabular}
}
\end{table}

\renewcommand{\arraystretch}{0.8}
\begin{table}[H]
\centering
\resizebox{\textwidth}{!}{
\begin{tabular}{ccccc}
\multicolumn{2}{l}{finish-5}& \multicolumn{1}{l}{}  & \multicolumn{2}{l}{finish-6}\\ \cline{1-2} \cline{4-5} 
\multicolumn{2}{|c|}{\begin{tabular}[c]{@{}c@{}}
“We finished the story…”\\(dem/KBW, sentence 17074)
\end{tabular}}      & \multicolumn{1}{c|}{} & \multicolumn{2}{c|}{\begin{tabular}[c]{@{}c@{}}
“…and finish the mortgage earlier.”\\(dem/KB7, sentence 3736)
\end{tabular}} \\ \cline{1-2} \cline{4-5} 
\multicolumn{1}{|c}{\cellcolor[HTML]{BBDAFF}Paraphrase candidate} & \multicolumn{1}{c|}{\cellcolor[HTML]{BBDAFF}Confidence}        & \multicolumn{1}{c|}{} & \multicolumn{1}{c}{\cellcolor[HTML]{BBDAFF}Paraphrase candidate}  & \multicolumn{1}{c|}{\cellcolor[HTML]{BBDAFF}Confidence}         \\ 
\multicolumn{1}{|c}{
Enjoy the story.}& \multicolumn{1}{c|}{\cellcolor[HTML]{9AFF99}0.42942}& \multicolumn{1}{c|}{} & \multicolumn{1}{c}{
Pay the mortgage.}& \multicolumn{1}{c|}{\cellcolor[HTML]{9AFF99}0.60147}\\\multicolumn{1}{|c}{
Hear the story.}& \multicolumn{1}{c|}{\cellcolor[HTML]{EFEFEF}0.39092}& \multicolumn{1}{c|}{} & \multicolumn{1}{c}{
Clear the mortgage.}& \multicolumn{1}{c|}{\cellcolor[HTML]{EFEFEF}\textbf{*0.47388}}\\\multicolumn{1}{|c}{
Read the story.}& \multicolumn{1}{c|}{\cellcolor[HTML]{EFEFEF}\textbf{*0.38208}}& \multicolumn{1}{c|}{} & \multicolumn{1}{c}{
Wait for the mortgage.}& \multicolumn{1}{c|}{\cellcolor[HTML]{EFEFEF}0.41958}\\\multicolumn{1}{|c}{
Tell the story.}& \multicolumn{1}{c|}{\cellcolor[HTML]{EFEFEF}0.37544}& \multicolumn{1}{c|}{} & \multicolumn{1}{c}{
Afford the mortgage.}& \multicolumn{1}{c|}{\cellcolor[HTML]{EFEFEF}0.41605}\\\cline{1-2} \cline{4-5} 
\end{tabular}
}
\end{table}

\renewcommand{\arraystretch}{0.8}
\begin{table}[H]
\centering
\resizebox{\textwidth}{!}{
\begin{tabular}{ccccc}
\multicolumn{2}{l}{finish-7}& \multicolumn{1}{l}{}  & \multicolumn{2}{l}{finish-8}\\ \cline{1-2} \cline{4-5} 
\multicolumn{2}{|c|}{\begin{tabular}[c]{@{}c@{}}
“Adam had finished the list\\of instructions…”\\(fic/G0L, sentence 1487)
\end{tabular}}      & \multicolumn{1}{c|}{} & \multicolumn{2}{c|}{\begin{tabular}[c]{@{}c@{}}
“…finished the game with only…”\\(news/CH3, sentence 6700)
\end{tabular}} \\ \cline{1-2} \cline{4-5} 
\multicolumn{1}{|c}{\cellcolor[HTML]{BBDAFF}Paraphrase candidate} & \multicolumn{1}{c|}{\cellcolor[HTML]{BBDAFF}Confidence}        & \multicolumn{1}{c|}{} & \multicolumn{1}{c}{\cellcolor[HTML]{BBDAFF}Paraphrase candidate}  & \multicolumn{1}{c|}{\cellcolor[HTML]{BBDAFF}Confidence}         \\ 
\multicolumn{1}{|c}{
Reach (for) the list.}& \multicolumn{1}{c|}{\cellcolor[HTML]{EFEFEF}0.4608}& \multicolumn{1}{c|}{} & \multicolumn{1}{c}{
Play the game.}& \multicolumn{1}{c|}{\cellcolor[HTML]{9AFF99}0.56537}\\\multicolumn{1}{|c}{
Look (at) the list.}& \multicolumn{1}{c|}{\cellcolor[HTML]{EFEFEF}0.42192}& \multicolumn{1}{c|}{} & \multicolumn{1}{c}{
Miss the game.}& \multicolumn{1}{c|}{\cellcolor[HTML]{9AFF99}\textbf{*0.50261}}\\\multicolumn{1}{|c}{
Read the list.}& \multicolumn{1}{c|}{\cellcolor[HTML]{EFEFEF}\textbf{*0.38076}}& \multicolumn{1}{c|}{} & \multicolumn{1}{c}{
Save the game.}& \multicolumn{1}{c|}{\cellcolor[HTML]{EFEFEF}0.48052}\\\multicolumn{1}{|c}{
Write the list.}& \multicolumn{1}{c|}{\cellcolor[HTML]{EFEFEF}0.42559}& \multicolumn{1}{c|}{} & \multicolumn{1}{c}{
Watch the game.}& \multicolumn{1}{c|}{\cellcolor[HTML]{EFEFEF}0.40168}\\\multicolumn{1}{|c}{
Develop the list.}& \multicolumn{1}{c|}{\cellcolor[HTML]{EFEFEF}0.35096}& \multicolumn{1}{c|}{} & \multicolumn{1}{c}{
Control the game.}& \multicolumn{1}{c|}{\cellcolor[HTML]{EFEFEF}0.28200}\\\cline{1-2} \cline{4-5} 
\end{tabular}
}
\end{table}

\renewcommand{\arraystretch}{0.8}
\begin{table}[H]
\centering
\resizebox{\textwidth}{!}{
\begin{tabular}{ccccc}
\multicolumn{2}{l}{finish-9}& \multicolumn{1}{l}{}  & \multicolumn{2}{l}{finish-10}\\ \cline{1-2} \cline{4-5} 
\multicolumn{2}{|c|}{\begin{tabular}[c]{@{}c@{}}
“She finished the whisky.”\\(fic/K8V, sentence 3338)
\end{tabular}}      & \multicolumn{1}{c|}{} & \multicolumn{2}{c|}{\begin{tabular}[c]{@{}c@{}}
“...I want to stay on here to finish the job.”\\(news/CH3, sentence 307)
\end{tabular}} \\ \cline{1-2} \cline{4-5} 
\multicolumn{1}{|c}{\cellcolor[HTML]{BBDAFF}Paraphrase candidate} & \multicolumn{1}{c|}{\cellcolor[HTML]{BBDAFF}Confidence}        & \multicolumn{1}{c|}{} & \multicolumn{1}{c}{\cellcolor[HTML]{BBDAFF}Paraphrase candidate}  & \multicolumn{1}{c|}{\cellcolor[HTML]{BBDAFF}Confidence}         \\ 
\multicolumn{1}{|c}{
Drink the whisky.}& \multicolumn{1}{c|}{\cellcolor[HTML]{9AFF99}0.58234}& \multicolumn{1}{c|}{} & \multicolumn{1}{c}{
Do the job.}& \multicolumn{1}{c|}{\cellcolor[HTML]{9AFF99}0.58553}\\\multicolumn{1}{|c}{
Gulp the whisky.}& \multicolumn{1}{c|}{\cellcolor[HTML]{EFEFEF}\textbf{*0.41849}}& \multicolumn{1}{c|}{} & \multicolumn{1}{c}{
Get the job.}& \multicolumn{1}{c|}{\cellcolor[HTML]{EFEFEF}\textbf{*0.48158}}\\\multicolumn{1}{|c}{
Look (at) the whisky.}& \multicolumn{1}{c|}{\cellcolor[HTML]{EFEFEF}0.41755}& \multicolumn{1}{c|}{} & \multicolumn{1}{c}{
Find the job.}& \multicolumn{1}{c|}{\cellcolor[HTML]{EFEFEF}0.42892}\\\multicolumn{1}{|c}{
Toss back the whisky.}& \multicolumn{1}{c|}{\cellcolor[HTML]{EFEFEF}\textbf{*0.38207}}& \multicolumn{1}{c|}{} & \multicolumn{1}{c}{
Have the job.}& \multicolumn{1}{c|}{\cellcolor[HTML]{EFEFEF}0.37238}\\\cline{1-2} \cline{4-5} 
\end{tabular}
}
\end{table}

\renewcommand{\arraystretch}{0.8}
\begin{table}[H]

\begin{tabular}{ccccc}
\multicolumn{2}{l}{finish-11}& \multicolumn{1}{l}{}  & \multicolumn{2}{l}{}\\ \cline{1-2}
\multicolumn{2}{|c|}{\begin{tabular}[c]{@{}c@{}}
“Finish the last packet of cigarettes…”\\(news/BM4, sentence 1431)
\end{tabular}}      & \multicolumn{1}{c}{} & \multicolumn{2}{c}{\begin{tabular}[c]{@{}c@{}}
\end{tabular}} \\ \cline{1-2}
\multicolumn{1}{|c}{\cellcolor[HTML]{BBDAFF}Paraphrase candidate} & \multicolumn{1}{c|}{\cellcolor[HTML]{BBDAFF}Confidence}        & \multicolumn{1}{c}{} & \multicolumn{1}{c}{}  & \multicolumn{1}{c}{}\\ 
\multicolumn{1}{|c}{
Carry the packet.}& \multicolumn{1}{c|}{\cellcolor[HTML]{EFEFEF}0.44237}& \multicolumn{1}{c}{} & \multicolumn{1}{c}{}& \multicolumn{1}{c}{}\\\multicolumn{1}{|c}{
Crumple the packet.}& \multicolumn{1}{c|}{\cellcolor[HTML]{EFEFEF}0.40580}& \multicolumn{1}{c}{} & \multicolumn{1}{c}{}& \multicolumn{1}{c}{}\\\multicolumn{1}{|c}{
Smoke the packet.}& \multicolumn{1}{c|}{\cellcolor[HTML]{EFEFEF}\textbf{*0.35518}}& \multicolumn{1}{c}{} & \multicolumn{1}{c}{}& \multicolumn{1}{c}{}\\\multicolumn{1}{|c}{
Open the packet.}& \multicolumn{1}{c|}{\cellcolor[HTML]{EFEFEF}0.36162}& \multicolumn{1}{c}{} & \multicolumn{1}{c}{}& \multicolumn{1}{c}{}\\\cline{1-2} 
\end{tabular}

\end{table}
\vspace{2mm}

\section{Colophon}

Servers hosted on Amazon Web Services were used to process data. The specifics of how these servers were configured can be found in Appendix III. These servers ran Ubuntu and scripts were written in Python 3.6, \texttt{https://www.python.org}. The experiment made extensive use of the NLTK (to parse the BNC) and gensim (Rehurek \& Sojka 2010) modules. The updated Stanford dependency parser was also used (Manning et al. 2016; Manning \& Sch\"{u}ster 2016). Locally, I used Sublime Text 3 as a Python IDE and text editor. PuTTY was my SSH client of choice and FileZilla was used to upload large files (such as the BNC) to the servers’ Elastic Block Store. The code for this paper is available as \texttt{doi:10.5281/zenodo.569505} and on GitHub:\hfill\break \texttt{https://github.com/albertomh/ug-dissertation}.\hfill\break\vspace{5mm}

\section{AWS server architecture}

The experiment was carried out on AWS servers running Ubuntu 16.04 (Long Term Support version). The creation of word vectors and evaluation of cosine similarity between targets and candidates required a t2.large instance. word2vec stores parameters as arrays of `vocabulary size * size of floats' (4 bytes). Three matrices of these characteristics are kept in RAM at any one point. In this experiment a matrix of 10 000 * 100 words was kept unfragmented in memory. This required a server with at least 4GB of RAM. For scraping VNP collocates from the BNC a t2.micro instance was used. Both the large and micro servers use Intel Xeon processors, which provide a balance of computational, memory and network resources and are burstable beyond baseline performance on demand. The BNC was kept on the Elastic Block Store chunked in XML files. Scripts ran concurrently with the EBS instance to access the BNC. Scraped data was stored as Python data types (lists, dictionaries) in text files and then interpreted using the ast module when needed (so as to not have them permanently loaded in memory).\hfill\break\vspace{2mm}

\begin{figure}[H]
\centering
\includegraphics[width=0.8\textwidth]{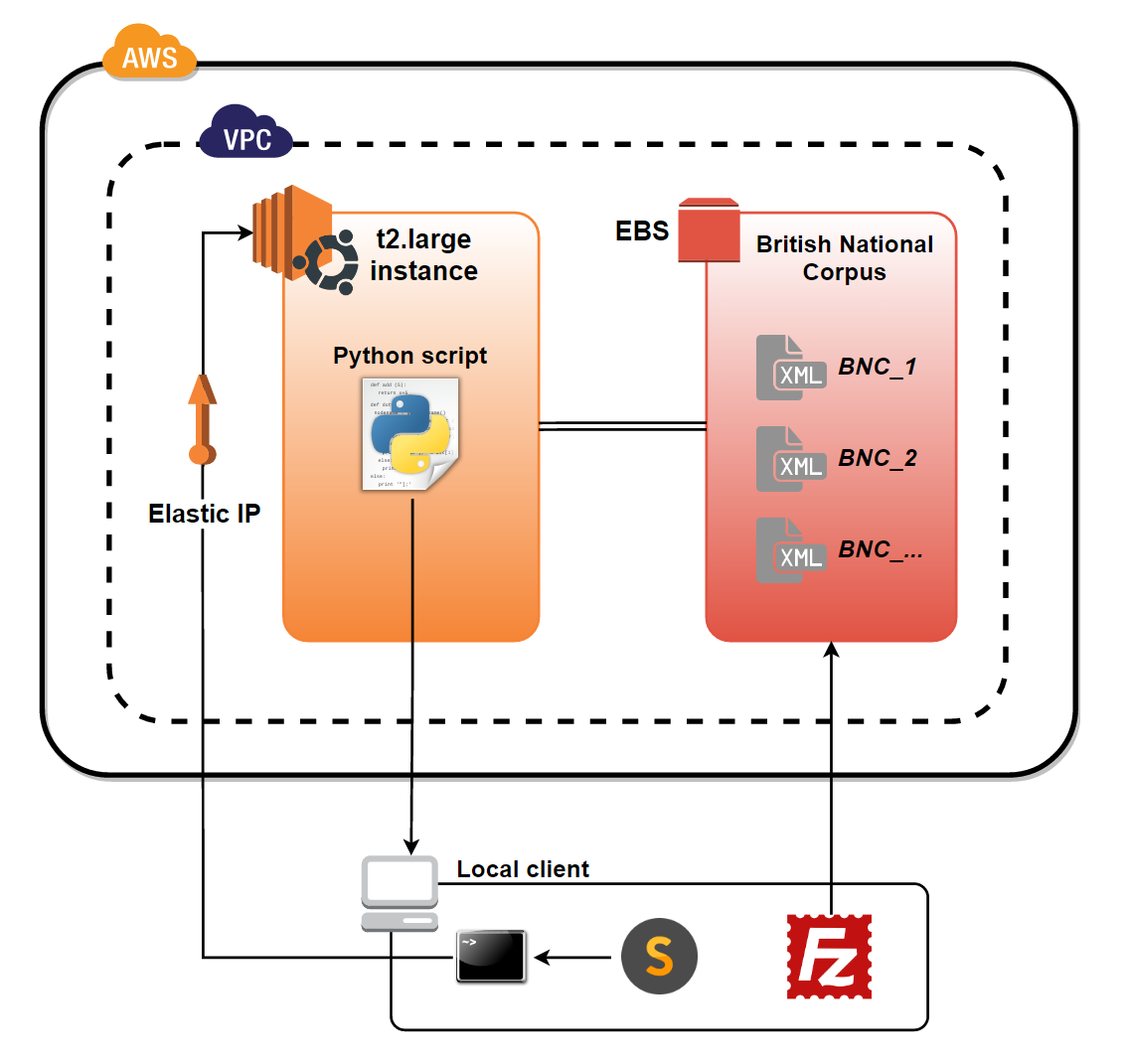}
\captionsetup{width=0.9\linewidth, labelformat=empty}
\caption[]{The infrastructure for the experiments was built using Amazon Web \mbox{Services}. Elastic Compute Cloud servers processed data held in Elastic Block Stores.}
\end{figure}

\end{document}